\documentclass{article}
\usepackage{ijcai24}

\usepackage{times}
\usepackage{soul}
\usepackage{url}
\usepackage[hidelinks]{hyperref}
\usepackage[utf8]{inputenc}
\usepackage[small]{caption}
\usepackage{graphicx}
\usepackage{amsmath}
\usepackage{amsthm}
\usepackage{booktabs}
\usepackage{algorithm}
\usepackage{algorithmicx}
\usepackage{algpseudocode}

\usepackage[switch]{lineno}

\usepackage{multirow}
\usepackage{amssymb}
\usepackage{color}
\usepackage{subcaption}

\algtext*{EndFor}
\algtext*{EndIf}
\algtext*{EndWhile}

\def \minus#1{{($- #1$)}}
\def \plus#1{{($+ #1$)}}

\newtheorem{definition}{Definition}

\title{Boundary Matters: A Bi-Level Active Finetuning Framework}

\author{
Han Lu$^{1}$ \and
Yichen Xie$^{2}$ \and
Xiaokang Yang$^{1}$ \and
Junchi Yan$^{1,\dag}$\\
\affiliations
$^{1}$ Department of Computer Science and Engineering, Shanghai Jiao Tong University \\ 
$^{2}$ University of California, Berkeley \\
$^\dag$ Correspondence author\\
}

\begin{document}

\maketitle

\begin{abstract}
The pretraining-finetuning paradigm has gained widespread adoption in vision tasks and other fields, yet it faces the significant challenge of high sample annotation costs. To mitigate this, the concept of active finetuning has emerged, aiming to select the most appropriate samples for model finetuning within a limited budget. Traditional active learning methods often struggle in this setting due to their inherent bias in batch selection. Furthermore, the recent active finetuning approach has primarily concentrated on aligning the distribution of selected subsets with the overall data pool, focusing solely on diversity. In this paper, we propose a Bi-Level Active Finetuning framework to select the samples for annotation in one shot, which includes two stages: core sample selection for diversity, and boundary sample selection for uncertainty. The process begins with the identification of pseudo-class centers, followed by an innovative denoising method and an iterative strategy for boundary sample selection in the high-dimensional feature space, all without relying on ground-truth labels. Our comprehensive experiments provide both qualitative and quantitative evidence of our method's efficacy, outperforming all the existing baselines.
\end{abstract}

\section{Introduction}
The advancement of deep learning significantly depends on extensive training data. However, it is challenging to annotate large-scale datasets, which requires significant human labor and resources. To mitigate this, the pretraining-finetuning paradigm has gained widespread adoption. In this paradigm, models are first pretrained in an unsupervised manner on large datasets, then finetuned on a smaller, labeled subset. While there is substantial research about both unsupervised pretraining~\cite{devlin2018bert,grill2020bootstrap,he2021masked,huang2021spatio} and supervised finetuning~\cite{hu2021lora,liu2021p}, the optimization of sample set selection for annotation has received less attention, especially under limited labeling resources.

Despite the crucial importance, traditional active learning methods ~\cite{sener2017active,sinha2019variational,parvaneh2022active} excel at selecting relevant samples from scratch but face challenges integrating into the pretraining-finetuning framework. This issue is highlighted in the literature~\cite{bengar2021reducing,xie2023active}. The central problem lies in the batch-selection approach; while it is effective for initial training, it becomes less suitable in the pretraining-finetuning paradigm, primarily due to constrained annotation budgets and inherent biases in small batch selections.

\begin{figure}[tb!]
    \centering
    \includegraphics[width=\linewidth]{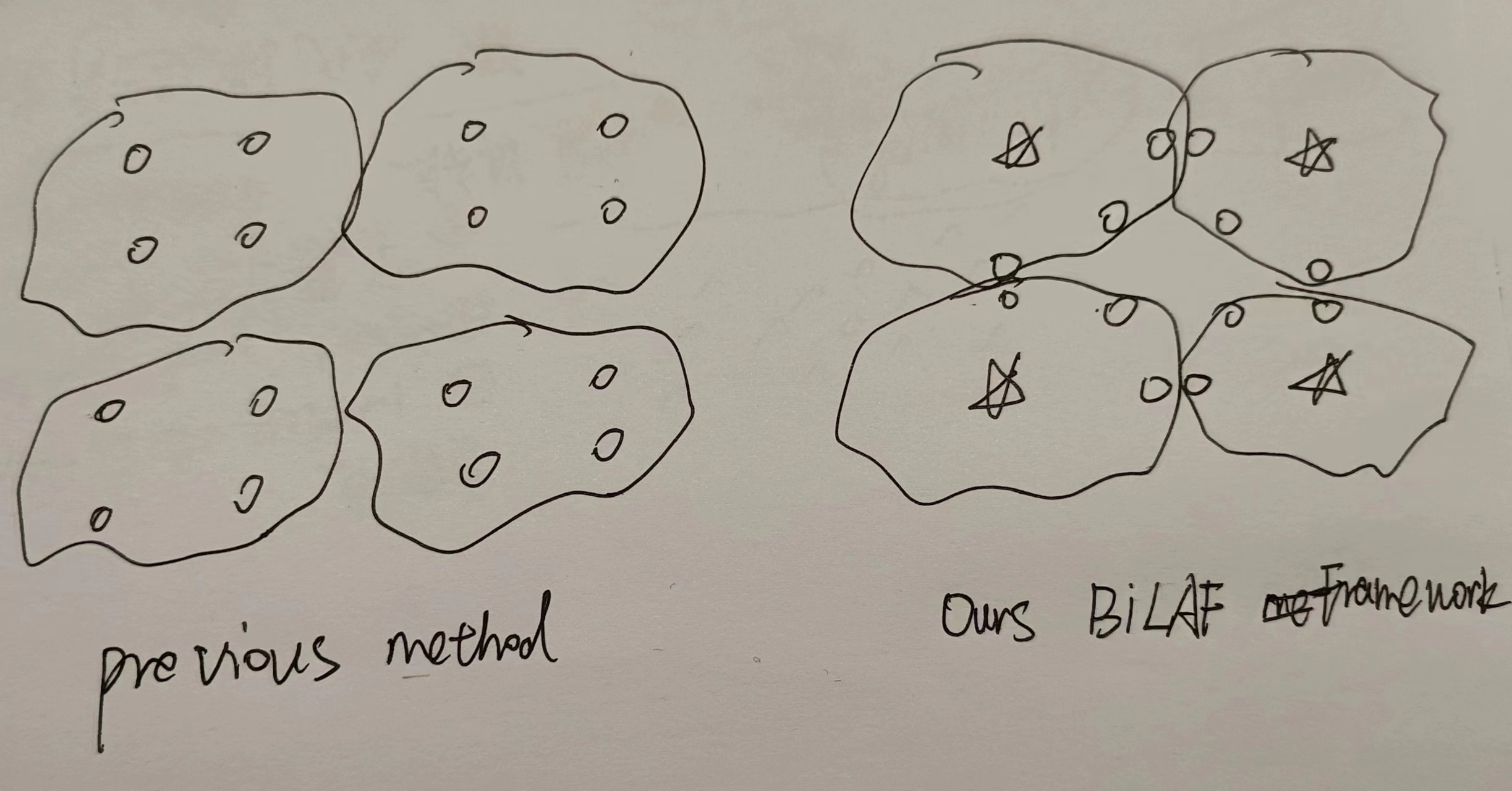}
    \caption{\textbf{Design philosophy of our BiLAF framework.} In contrast to previous methods, our method ensures the selection of central samples to maintain diversity while also reserving capacity to choose boundary samples to enhance decision boundary learning. }
    \label{fig:intro}
\end{figure}

To fill in the research gap, the \textit{Active Finetuning} task has been formulated in \cite{xie2023active}, which focuses on the selection of samples for supervised finetuning using pretrained models.
This method optimizes sample selection by minimizing the distributional gap between the selected subset and the entire data pool.
Despite its notable performance, it fundamentally focuses on the diversity of samples, which overlooks the crucial aspect of uncertainty in data selection. Particularly, as the volume of data increases, the selected samples become increasingly redundant, thereby limiting the model's capabilities, which is corroborated by the empirical experiments.

To address its shortcomings, we propose an innovative \textbf{Bi-Level Active Finetuning Framework (BiLAF)}, designed to ensure that the sample selection process  effectively balances both diversity and uncertainty, as shown in  Fig.~\ref{fig:intro}. The primary challenge lies in measuring uncertainty within the pretraining-finetuning paradigm. Traditional strategies, such as relying on classification confidence, are inapplicable in this scenario due to the lack of iterative training based on class labels. Alternatively, our focus shifts to the model's decision boundary, a concept that has extensive theoretical~\cite{lei2023understanding} and methodological research~\cite{burges1998tutorial,cao2019learning} across various domains. Intriguingly, the global feature space of pretrained models can represent the interrelations between samples from different classes, allowing to find those support samples close to the decision boundary.

To elaborate, our BiLAF framework operates in two distinct stages. The first stage, \textit{Core Samples Selection}, is dedicated to selecting central samples of each class. The selection method can be any that optimizes center samples, such as K-Means or ActiveFT~\cite{xie2023active}. 
The second stage, \textit{Boundary Samples Selection}, involves our novel unsupervised denoising method to pinpoint outliers. Following this, we systematically select boundary samples around each center, using our specially developed boundary score, while avoiding redundancy by excluding similar samples.


Our contributions are summarized as follows: 
\begin{itemize}
    \item We propose a Bi-Level Active Finetuning Framework (BiLAF) emphasizing boundary importance to balance sample diversity and uncertainty.
    \item We innovate an unsupervised denoising method to eliminate the outlier samples and an iterative strategy to efficiently identifying boundary samples in feature space.
    \item Extensive experiments and ablation studies demonstrate the effectiveness of our method. Compared to current state-of-the-art approach, our method achieves a remarkable improvement of nearly 3\% on CIFAR100 and approximately 1\% on ImageNet.
\end{itemize}

\section{Related Work}
\subsection{Active Learning / Finetuning}
Active learning strives to maximize the utility of a limited annotation budget by selecting the most valuable data samples. In most pool-based scenarios, existing algorithms typically employ criteria of uncertainty or diversity for sample selection.  Uncertainty-based methods typically make choices based on the difficulty of each sample, which is measured by various heuristics such as posterior probability~\cite{lewis1994heterogeneous,wang2016cost}, entropy~\cite{joshi2009multi,luo2013latent}, loss function~\cite{huang2021semi,yoo2019learning}, or the influence on the model performance~\cite{freytag2014selecting,liu2021influence}. In contrast, diversity-based algorithm selects representative data samples to approximate the distribution of the original data pool. The diversity can be quantified by the distance between the global~\cite{sener2017active} and local~\cite{agarwal2020contextual,xie2023towards} representations or other metrics like gradient directions~\cite{ash2019deep} or adversarial loss~\cite{kim2021task,sinha2019variational}.

However, the algorithms mentioned above are primarily designed for from-scratch training and face challenges within the pretraining-finetuning paradigm~\cite{hacohen2022active,xie2023active}. In response, ActiveFT~\cite{xie2023active} has been specifically developed for this context. It selects data by bringing the distribution of selected samples close to the original unlabeled pool in the feature space. ActiveFT tends to favor high-density areas in its selection, often overlooking boundary samples. In contrast, our proposed algorithm incorporates both diversity and uncertainty into the selection process, which is important for determining the decision boundary in the supervised finetuning.

\subsection{Decision Boundaries in Neural Networks.}
Decision boundaries are crucial in neural network-based classification models, significantly influencing performance and interpretability~\cite{lei2023understanding,li2023support}. Their optimization can greatly enhance model generalizability and accuracy, particularly in complex, high-dimensional data spaces~\cite{xu2019locally,cao2019learning}. In Support Vector Machines (SVMs), the decision boundary is fundamental to separating hyperplanes, and maximizing the geometric margin around this boundary is key for robust classification~\cite{burges1998tutorial}. This concept is also applicable to neural networks, where similar margin maximization can improve generalization. In the context of imbalanced datasets, adjusting the decision boundary is essential for the accurate classification of minority classes. Techniques like Label-Distribution-Aware Margin (LDAM) loss~\cite{cao2019learning} and Enlarged Large Margin (ELM) loss~\cite{kato2023enlarged} have been developed to modify the decision boundary, thus balancing generalization errors across classes. It has been observed that neural networks typically rely on the most discriminative or simplest features to construct decision boundaries~\cite{ortiz2020hold,shah2020pitfalls}. A theoretical framework for assessing decision boundary complexity through the novel metric of decision boundary (DB) variability, and its inverse relationship with generalizability, is presented in~\cite{lei2023understanding}.

However, in the field of active finetuning, the selection of Decision Boundaries has not yet been a focus of attention, although features from the pretrained model can provide an accurate representation of the sample boundary relationships. For the first time in the active finetuning task, we introduce a focus on Boundary Sample analysis.

\begin{figure*}[th!]
    \centering
    \includegraphics[width=0.9\linewidth]{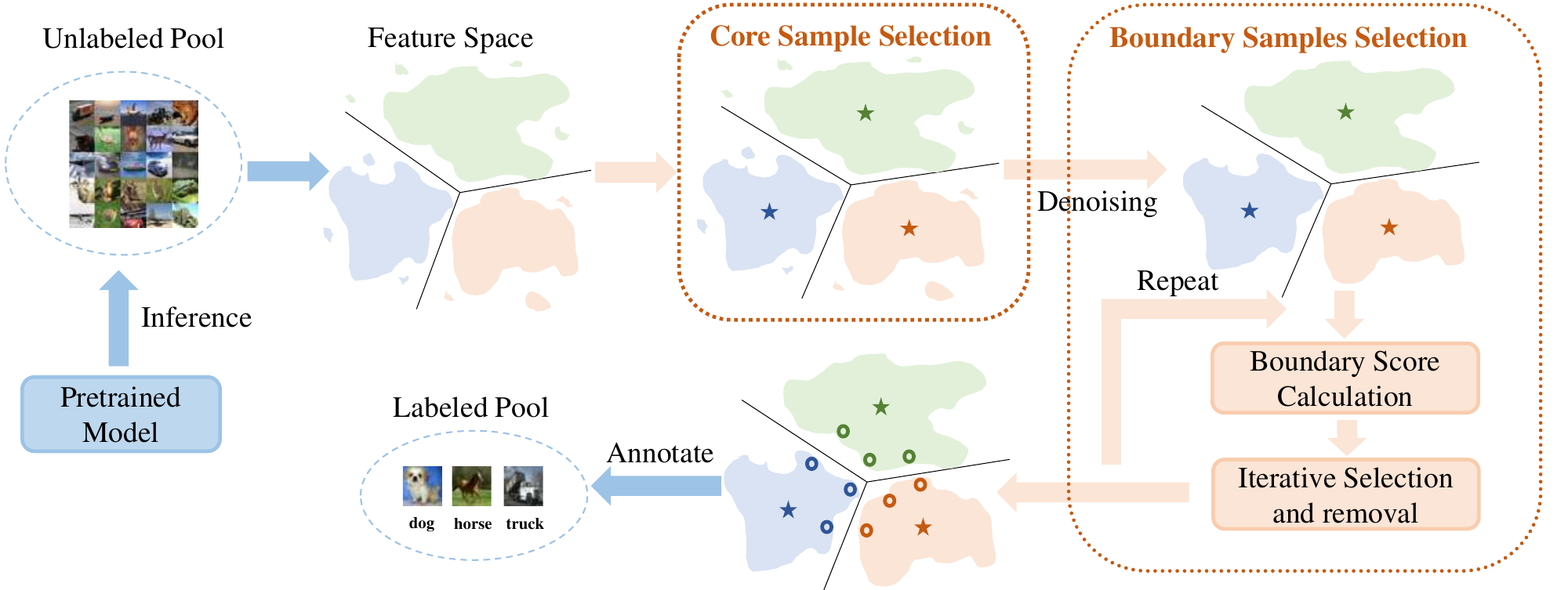}
    \caption{\textbf{Our BiLAF framework in the Active Finetuning task.} Within the high-dimensional feature space, the first stage core sample selection is dedicated to identifying pseudo-class centers. Building upon this, we have developed a denoising method to filter out noise samples. Subsequently, we compute the Boundary Score metric for each sample, which aids in the iterative selection of samples and the removal of candidates from the pool. Ultimately, the selected samples are labeled for supervised finetuning. }
    \label{fig:framework}
\end{figure*}

 
\section{BiLAF: Bi-Level Active Finetuning}

In this section, we introduce our novel \textbf{Bi}-\textbf{L}evel \textbf{A}ctive \textbf{F}inetuning Framework (\textbf{BiLAF}), as illustrated in Fig.~\ref{fig:framework}. BiLAF operates in two distinct stages: Initially, the \textit{Core Samples Selection} stage is employed (Sec.~\ref{sec:core_samples}). We identify multiple pseudo-class centers, intentionally selecting a slightly larger number of centers than the actual number of classes to ensure extensive coverage across all classes and to account for multiple centers within individual classes. The second stage involves our \textit{Boundary Samples Selection method} (Sec.~\ref{sec:boundart_selction}), which focuses on accurately identifying boundary samples for each class.  Algorithm~\ref{alg:BiLAF} summarizes the complete workflow of BiLAF.

\subsection{Preliminary: Active Finetuning Task}
\label{sec:formulation}

The active finetuning task is formulated in~\cite{xie2023active}. Apart from a large unlabeled data pool $\mathcal{P}^u=\{\mathbf{x}_i\}_{i\in [N]}$ where $[N]=\{1,2,\dots,N\}$ same in the traditional active learning task, we have the access to a deep neural network model $f(\cdot;w_0)$ with the pretrained weight $w_0$.  It should be noted that the function $f(\cdot;w_0)$ can be pretrained either on the current data pool $\mathcal{P}^u$ or any other data sources.
Using the pretrained model, we can map the data sample $\mathbf{x}_i$ to the high dimensional feature space as $\mathbf{f}_i=f(\mathbf{x}_i;w_0) \in \mathbb{R}^d$, where $\mathbf{f}_i$ is the \textit{normalized} feature of $\mathbf{x}_i$. Therefore, we can derive the feature pool $\mathcal{F}^u=\{\mathbf{f}_i\}_{i\in[N]}$ from $\mathcal{P}^u$. 

Our task is to select the subset $\mathcal{P}^u_{\mathcal{S}}$ from $\mathcal{P}^u$ for annotation and the following supervised finetuning.  $\mathcal{S}=\{s_j\in[N]\}_{j\in[B]}$ is a sampling strategy so that $\mathcal{P}_{\mathcal{S}}^u=\{\mathbf{x}_{s_j}\}_{j\in[B]}$, where $B$ is the annotation budget for supervised finetuning. The model would have access to the labels $\{\mathbf{y}_{s_j}\}_{j\in[B]}\subset\mathcal{Y}$ of this subset through the oracle, obtaining a labeled data pool $\mathcal{P}^l_\mathcal{S}=\{\mathbf{x}_{s_j},\mathbf{y}_{s_j}\}_{j\in[B]}$, where $\mathcal{Y}$ is the label space. Then, the pretrained model $f(\cdot;w_0)$ is finetuned on $\mathcal{P}^l_\mathcal{S}$ supervisedly.
Our objective is to choose the optimal sampling strategy to select the labeled set under the given budget to minimize the expectation error of the finetuned model.

\subsection{Core Samples Selection} 
\label{sec:core_samples}
Core Samples Selection is a fundamental approach in active learning, aimed at identifying key samples that represent the entire  dataset. Popular techniques include K-Means, Coreset~\cite{sener2017active}, and ActiveFT~\cite{xie2023active}, which primarily differ in their target designs and optimization processes. Within this framework, we adopt ActiveFT~\cite{xie2023active}, the most advanced model to date, as our primary method for the initial phase. For specific details on implementation of ActiveFT, please refer to Appendix~\ref{app:app_active}. We utilize ActiveFT to obtain $K$ core samples, serving as pseudo-class centers, where $K$ is the predefined budget for central samples.
Then, we proceed with the boundary sample selection based on these $K$ pseudo-class centers.

\subsection{Boundary Samples Selection}
\label{sec:boundart_selction}
Decision boundaries are critical in classification problems, particularly in Active Finetuning.
Selecting boundary samples for each class effectively can greatly enhance the training of the model's decision boundaries, even within a limited budget, thus improving overall performance. 

By leveraging pretrained models, data samples can be mapped to robust feature representations, which effectively elucidates the relationships between each sample, its intra-class counterparts, and inter-class samples from diverse classes. We design an innovative method for boundary sample selection for the first time in this task. Given the pseudo-class centers, our method comprises three steps: \textit{1) Identifying the samples that belong to each pseudo-class center. 2) Implementing denoising processes on these samples. 3) Developing metrics to precisely select boundary points for each pseudo-class}.

\subsubsection{Step1: Clustering Process}

Given $K$ pseudo-class centers denoted by $C = \{c_1, c_2, \ldots, c_K\}$ where $c_i \in [N]$, for the sample $\mathbf{x}_j$ with its corresponding feature $\mathbf{f}_j \in \mathbb{R}^d$, it is assigned to the pseudo-class center $c_i$ that minimizes the distance $D(\mathbf{f}_j, \mathbf{f}_{c_i})$, which can be represented as:
\begin{equation}
   c_i = \arg \min_{c \in C} D(\mathbf{f}_j, \mathbf{f}_{c}) 
\end{equation}
where $D(\cdot, \cdot)$ is the distance function. 
The choice of $D$ can be any distance metric, depending on the clustering design of the center selection model, which inherently assign each sample point to a corresponding pseudo-class center during optimization. Therefore, The Euclidean distance is used in our implementation.

\subsubsection{Step2: Denoising process.} 
Boundary samples within a pseudo-class provide critical insights for optimizing the decision boundary. However, they can also introduce noise, potentially hindering the model's performance.
For each class $i$ containing $N_i$ samples,  we set a removal ratio $P_{rm}$ to identify and eliminate  $N_{i,rm} = N_i \cdot P_{rm}$ peripheral noisy samples from the candidate boundary set. This elimination is based on the sample density in the feature space, where we define the density distance as follows.
\begin{definition}[Density Distance]
    The density distance of a point is defined as the average distance to its $k$ nearest neighbors. Formally, for a given sample $x_j$ with the feature $\mathbf{f}_j$, its density distance $ \rho({\mathbf{x}_j})$ is defined as:
    \begin{equation}
     \rho(\mathbf{x}_j) = \frac{1}{k} \sum_{l=1}^{k} D(\mathbf{f}_j, \mathbf{f}_{n_{jl}})
     \label{equ:density}
    \end{equation}
    where $n_{jl}$ is the $l$-th nearest  neighbor index of the sample $x_j$ in the feature space and $D$ is a distance function.
\end{definition}

We get inspiration from the classical clustering method DBSCAN~\cite{ester1996density} to propose an Iterative Density-based Clustering (IDC) algorithm. IDC clusters the candidate points $\mathbf{X}_i = \{ \mathbf{x}_{a_{i,1}}, \mathbf{x}_{a_{i,2}}, ... \mathbf{x}_{a_{i,N_i}} \}$ where $a_{i,j} \in [N]$, beginning with the core sample index as the initialization set $U_i = \{c_i\}$. In the following each iteration, a fraction $P_{in}$ of points will be included, which means $N_{i, in} = N_i \cdot P_{in}$ peripheral samples are absorbed into the existing cluster $U_{i}$.

The density distance of each candidate point is redefined from Eq.~\ref{equ:density} as the average distance to the nearest $ k $ selected points which are in $U_{i}$. Formally, the density distance $\rho(\mathbf{x}_j)$ for each remaining point $\mathbf{x}_{j} $ is calculated as:
\begin{equation}
 \rho(\mathbf{x}_j) = \frac{1}{k} \sum_{l=1, {n_{jl}} \in U_i}^{k} D(\mathbf{f}_j, \mathbf{f}_{n_{jl}})   
 \label{equ:density_IDC}
\end{equation}
where $n_{jl}$ is the nearest $l$-th sample index of $\mathbf{x}_j$ which has been already included in the cluster. The $N_{i,in}$ densest 
samples with lowest density distance are chosen in each iteration. The process repeats until all points are selected. The order of inclusion into the class indicates the point's proximity to the center, with points added later more likely to be noisy. Therefore, we can remove the last $N_{i, rm}$ samples.

\begin{algorithm}[tb!]
    \caption{\textbf{Pseudo-code for BiLAF}}
    \label{alg:BiLAF}
    \textbf{Input}: Unlabeled data pool $\mathcal{P}^u = \{\mathbf{x}_i\}_{i\in[N]}$, pretrained model $f(\cdot;w_0)$, annotation budget $B$.\\
    \textbf{Parameter}:  Core samples budget $K$, removal ratio $P_{rm}$, density neighbours $k$, IDC cluster ratio $P_{in}$, opponent penalty coefficient $\delta$, distance function $D(\cdot,\cdot)$.\\
    \textbf{Output}: The selected samples index $\mathcal{S} = \{s_j\}_{j\in[B]}$
    \begin{algorithmic}[1] 
        \Statex $\triangleright$ \textit{Stage 1: Core Samples Selection}
        \For{$i\in[N]$}
            \State $\mathbf{f}_i=f(\mathbf{x}_i;w_0)$
        \EndFor
        \State Select $K$ centers in feature space $\mathcal{F}$ using ActiveFT
        \Statex \Comment{\textit{Can be K-Means and any other core selection method}}
        \vspace{5pt}
        \Statex $\triangleright$ \textit{Stage 2: Boundary Samples Selection}
        \For{each pseudo-class $c_i$}
            \vspace{5pt}
            \Statex  $\quad \triangleright$ \textit{Step1: Assign the pseudo-class samples}
            \State Initial the candidate index set $ U^0_i = \{\}$
            \For{$j\in[N]$}
                \If{$c_i == \arg \min_{c \in C} D(\mathbf{f}_j, \mathbf{f}_{c})$}
                    \State Add the index $j$ to the set $U^0_i$
                \EndIf
            \EndFor
            
            \Statex $\quad \triangleright$
            \textit{Step2: Iterative Density-based cluster then denoise}

            \State Initialize the pseudo-class cluster $U_i = \{c_i\}$.
            \While{$|U^0_i| > |U_i|$}
                \For{each candidate index $j \in U^0_i$ and $j \notin U_i$}
                    \State $ \rho(\mathbf{x}_j) = \frac{1}{k} \sum_{l=1, \mathbf{n_{jl}} \in U_i}^{k} D(\mathbf{f}_j, \mathbf{f}_{n_{jl}}) $ where  
                    \Statex $\quad \quad \quad \quad \quad \quad n_{jl}$ is the nearest $l$-th sample index of $\mathbf{x}_j$.                    
                \EndFor
                \State Sort samples by increasing density distance $\rho(\mathbf{x}_j)$
                \State Add the previous $P_{in} \cdot |U^0_i|$  samples to  $U_i$
            \EndWhile
            \State Remove the last $P_{rm} \cdot |U_i|$ samples from $U_i$ to denoise
            
            \vspace{5pt}
            \Statex $\quad \triangleright$ \textit{Step3: Selection Process}

             \State Initialize opponent penalty counts $t_l = 0$ for $l \neq i$
            \For{$m=1$ to $B_i$}
                \For{each candidate index $j \in U_i$}              
                    \State $d_{intra}(\mathbf{x}_j)  = \frac{1}{|U_i|} \sum_{l \in U_i} D(\mathbf{f}_j, \mathbf{f}_l)$       
                    \State $BS(\mathbf{x}_j) = \min_{c_l \in C}^{i \neq l} \frac{\delta^{t_l} \cdot D(\mathbf{f}_j, \mathbf{f}_{c_l}) - d_{intra}(\mathbf{x}_j)}{\max(D(\mathbf{f}_j, \mathbf{f}_{c_l}), d_{intra}(\mathbf{x}_j))}$
                \EndFor
                \State Select  $\mathbf{x}_{j_{min}}$ with the lowest Boundary Score if
                \Statex $\quad \quad \quad \quad $ $m > 1$, otherwise set $j_{min} = c_i$.
                \State Add the index $j_{min}$ to selected samples set $\mathcal{S}$
                \State $t_l = t_l + 1$ where nearest opponent class is $l$ if 
                \Statex $\quad \quad \quad \quad $ $m > 1$.
                \State Remove the nearest $|U_i| / B_i$ samples of $\mathbf{x}_{j_{min}}$ 
                \Statex $\quad \quad \quad \quad $ from $U_i$
            \EndFor
        \EndFor
        \State \textbf{return} the selected samples index $\mathcal{S}$
    \end{algorithmic}    
\end{algorithm}

\subsubsection{Step3: Selection Process.} 
Having effectively removed the noise samples, we then move on to selecting boundary samples from the pseudo-class centers. To facilitate this, we introduce a series of definitions that will be used to calculate the boundary score for each sample.

\begin{definition}[Intra-class Distance]
    Let $ \mathbf{x}_j $ be a sample in the set $U_i$ of the pseudo-class $ c_i $. The intra-class distance for $ \mathbf{x}_j $ is the average distance from $ \mathbf{f}_j $ to all other samples within the same class set $U_i$. It can be formally defined as:
    \begin{equation}
    d_{intra}(\mathbf{x}_j)  = \frac{1}{|U_i|} \sum_{l \in U_i} D(\mathbf{f}_j, \mathbf{f}_l)
    \end{equation}
    where $ D $ is the distance function and $ |U_i| $ is the number of samples in pseudo-class $ c_i $.
\end{definition}
\begin{definition}[Inter-class Distance]
    The inter-class distance for a sample $ \mathbf{x}_j $ in the pseudo-class $c_i$ is defined as the distance from $ \mathbf{f}_j $ to the nearest center of other pseudo-classes. It is defined as:
    \begin{equation}
    d_{inter}(\mathbf{x}_j) = \min_{c_l \in C, i \neq l} D(\mathbf{f}_j, \mathbf{f}_{c_l})
    \end{equation}
    where $ D $ is the distance function, $ C = \{c_1, \ldots, c_K\} $ are the pseudo-class centers, and $ i \neq  l$ ensures that the pseudo-class $c_i$ of the sample $ \mathbf{x}_j $ is excluded from the calculation.
\end{definition}

\begin{definition}[Boundary Score]
    The Boundary Score for sample $ \mathbf{x}_j $ in psuedo-class $ c_i $ can be defined as a function of both intra-class and inter-class distances. It is defined as:
    \begin{equation}
    BS(\mathbf{x}_j) = \frac{d_{inter}(\mathbf{x}_j) - d_{intra}(\mathbf{x}_j)}{\max(d_{inter}(\mathbf{x}_j) , d_{intra}(\mathbf{x}_j))}
    \label{equ:Boundary_Score}
    \end{equation}
    where $ d_{intra} $ and $ d_{inter} $ are the intra-class and inter-class distances, respectively. A smaller Boundary Score indicates closer proximity to the boundary.
\end{definition}

For each pseudo-class $c_i$ with the samples set $U_i$, we allocate the sample selection budget proportionally to the number of samples in the pseudo-class. Formally, the budget $B_i$ for pseudo-class $c_i$ can be represented as:
$ B_i = B \cdot|U_i|\ / \sum_{j=1}^K |U_j| $
where $|U_i|$ denotes the number of samples in pseudo-class $c_i$.

Initially, one might consider simply selecting the top $B_i$ samples with the lowest boundary scores. However, this approach risks concentrating the selected samples in specific areas. To circumvent this, we employ an \textbf{iterative selection and removal} strategy, which involves progressively selecting and removing candidate samples. Moreover, to prevent multiple samples from clustering at the same pseudo-class boundary against an opposing class center, we introduce an \textbf{opponent penalty}. This penalty curtails the influence of already selected boundary samples against other classes, thereby promoting a more diverse selection across different boundaries.

\textbf{Iterative selection and removal} involves selecting the samples with the lowest Boundary Score in each round and then removing the nearest $|U_i| / B_i$ samples surrounding the selected sample. The process repeats for $B_i$ times beginning with the center samples.

\textbf{Opponent penalty} monitors  the relationship between the currently selected samples and the boundaries of other pseudo-classes, imposing a penalty on pseudo-classes that have already been selected. Specifically, for the selected pool of pseudo-class $c_i$, if boundary samples with an opposing pseudo-class $l \neq i$ have been selected $t_l$ times, the inter-class distance for the subsequent samples will be scaled  by a factor of $\delta^{t_l}$, where the opponent penalty coefficient $\delta$ is a hyperparameter greater than 1. This indicates that the more often a sample is selected in relation to pseudo-class $l$'s boundary, the higher its Boundary Score with that pseudo-class becomes, making it less likely to be chosen in future rounds. Consequently, we recalibrate the Boundary Score for each sample in each iteration, leading to a modification from Eq.~\ref{equ:Boundary_Score}:
\begin{equation}
    BS(\mathbf{x}_j) = \min_{c_l \in C, i \neq l} \frac{\delta^{t_l} \cdot D(\mathbf{f}_j, \mathbf{f}_{c_l}) - d_{intra}(\mathbf{x}_j)}{\max(D(\mathbf{f}_j, \mathbf{f}_{c_l}), d_{intra}(\mathbf{x}_j))}
    \label{eq:calibrated_score}
\end{equation}
where $D(\mathbf{f}_j, \mathbf{f}_{c_l})$ is the distance of the current sample $\mathbf{x}_j$ to the opponent pseudo-class $l$ and the $\delta^{t_l}$ is the related penalty. 

For each pseudo-category $i$, we would select the $B_i$ samples through an iterative selection and removal process, incorporating an opponent penalty. Finally, all selected samples from different pseudo-categories are combined as the selected subset $\mathcal{P}_{\mathcal{S}}^u$.


\section{Experiments}
Our approach is evaluated on three image classification benchmarks with varying sampling ratios detailed in Sec.~\ref{sec:exp_setup}, comparing its performance against various baselines and conventional active learning techniques in Sec.~\ref{sec:bench_result}. The qualitative visualization is demonstrated in Sec.\ref{sec:analysis} and the comprehensive quantitative ablation study is provided in Sec.~\ref{sec:ablation}. All experiments were conducted on two GeForce RTX 3090 (24GB) GPUs. Source code will be made publicly available.

\begin{table*}[th!]
\renewcommand{\arraystretch}{1.2}
\caption{\textbf{Benchmark Results.} Experiments are conducted on three popular datasets with different annotation ratios. We report the mean and standard deviation over three trials. Traditional active learning methods require random initial data to start, therefore we use ``-'' to represent. Our approach has demonstrated a significant competitive advantage across the majority of scenarios, affirming its effectiveness.} 
\label{tab:benchmarks}
\centering
\resizebox{\linewidth}{!}{
\begin{tabular}{l|ccc|cccc|ccc}
     \toprule[1.0pt]
     \multirow{2}{*}{\textbf{Methods}} & \multicolumn{3}{c}{\textbf{CIFAR10}} & \multicolumn{4}{c}{\textbf{CIFAR100}} & \multicolumn{3}{c}{\textbf{ImageNet}}\\
     & $0.5\%$ & $1\%$ & $2\%$ & $1\%$ & $2\%$ & $5\%$ & $10\%$ & $1\%$ & $2\%$ & $5\%$ \\
     \midrule[0.8pt]
     \textbf{Random} & 77.3$\pm$2.6 &82.2$\pm$1.9 &88.9$\pm$0.4 &14.9$\pm$1.9 &24.3$\pm$2.0 &50.8$\pm$3.4 &69.3$\pm$0.7 & 45.1$\pm$0.8 & 52.1$\pm$0.6  &64.3$\pm$0.3\\
     \textbf{FDS} & 64.5$\pm$1.5 & 73.2$\pm$1.2 & 81.4$\pm$0.7 & 8.1$\pm$0.6 & 12.8$\pm$0.3 & 16.9$\pm$1.4 & 52.3$\pm$1.9 & 26.7$\pm$0.6 & 43.1$\pm$0.4 & 55.5$\pm$0.1\\
     \textbf{K-Means} & 83.0$\pm$3.5 & 85.9$\pm$0.8 & 89.6$\pm$0.6& 17.6$\pm$1.1 & 31.9$\pm$0.1 & 42.4$\pm$1.0 & 70.7$\pm$0.3 & - & - & -\\
     \midrule[0.8pt]
     \textbf{CoreSet} ~\cite{sener2017active} & - & 81.6$\pm$0.3 & 88.4$\pm$0.2 & - & 30.6$\pm$0.4 & 48.3$\pm$0.5 & 62.9$\pm$0.6 & - & 52.7$\pm$0.4 & 61.7$\pm$0.2\\
     \textbf{VAAL} ~\cite{sinha2019variational} & -& 80.9$\pm$0.5 & 88.8$\pm$0.3 & - & 24.6$\pm$1.1 & 46.4$\pm$0.8 & 70.1$\pm$0.4& - & 54.7$\pm$0.5 & 64.0$\pm$0.3\\
     \textbf{LearnLoss} ~\cite{yoo2019learning} & -& 81.6$\pm$0.6 & 86.7$\pm$0.4 & - & 19.2$\pm$2.2 & 38.2$\pm$2.8 & 65.7$\pm$1.1& - & 54.3$\pm$0.6 & 63.2$\pm$0.4\\
     \textbf{TA-VAAL}~\cite{kim2021task}  & - & 82.6$\pm$0.4 & 88.7$\pm$0.2 & - & 34.7$\pm$0.7 & 46.4$\pm$1.1 & 66.8$\pm$0.5 & - & 55.0$\pm$0.4 & 64.3$\pm$0.2 \\
     \textbf{ALFA-Mix}~\cite{parvaneh2022active} & - & 83.4$\pm$0.3 & 89.6$\pm$0.2 & - & 35.3$\pm$0.8 & 50.4$\pm$0.9 & 69.9$\pm$0.6 & - & 55.3$\pm$0.3 & 64.5$\pm$0.2 \\
     \midrule[0.8pt]
    \textbf{ActiveFT}~\cite{xie2023active} & \textbf{85.0}$\pm$0.4 & 88.2$\pm$0.4 & 90.1$\pm$0.2 & 26.1$\pm$2.6 & 40.7$\pm$0.9 & 54.6$\pm$2.3 & 71.0$\pm$0.5 & 50.1$\pm$0.3 & 55.8$\pm$0.3 & 65.3$\pm$0.1\\
    \midrule[0.8pt]
    \textbf{BiLAF (ours)} & 81.0$\pm$1.2 & \textbf{89.2}$\pm$0.6 & \textbf{92.5}$\pm$0.4& \textbf{31.8}$\pm$1.6 & \textbf{43.5}$\pm$0.8 & \textbf{62.8}$\pm$1.2 & \textbf{73.7}$\pm$0.5 & \textbf{50.8}$\pm$0.4 & \textbf{56.9}$\pm$0.3 & \textbf{66.2}$\pm$0.2 \\
     \bottomrule[1.0pt]
\end{tabular}
}
\end{table*}
 
\subsection{Experiment Setup}
\label{sec:exp_setup}

\textbf{Datasets and Evaluation Metrics.} Our approach is evaluated using three widely recognized datasets: CIFAR10, CIFAR100 \cite{krizhevsky2009learning}, and ImageNet-1k \cite{russakovsky2015imagenet}. 
The raw training data serve as the unlabeled pool $\mathcal{P}^u$ from which selections are made. We employ the \textit{Top-1 Accuracy} metric for performance evaluation.

\textbf{Baselines.}  We compare our approach with three heuristic baselines Random, FDS and K-Means; four active learning methods CoreSet~\cite{sener2017active}, VAAL~\cite{sinha2019variational}, LearnLoss~\cite{yoo2019learning}, TA-VAAL~\cite{kim2021task}, and ALFA-Mix~\cite{parvaneh2022active}; and the well-designed active finetuning method ActiveFT~\cite{xie2023active}. We utilize the results reported in ~\cite{xie2023active}.
The detailed information is listed in Appendix~\ref{app:app_exp_setup}.




\textbf{Implementation Details.}  In line with the SOTA method ActiveFT~\cite{xie2023active}, we use DeiT-Small~ \cite{touvron2021training} model, pretrained using the DINO~\cite{caron2021emerging} framework on ImageNet-1k in the unsupervised pretraining phase, due to its recognized efficiency and popularity.
For all three datasets, we resize images to $224\times224$ consistent with the pretraining for both data selection and supervised finetuning. 
In the core samples selection stage, we utilize ActiveFT and optimize the parameters $\theta_{\mathcal{S}}$ using the Adam \cite{kingma2014adam} optimizer (learning rate 1e-3) until convergence. We set the core number $K$ as $50(0.1\%), 250(0.5\%), 6405(0.5\%)$  for CIFAR10, CIFAR100 and ImageNet separately. In the boundary samples selection stage, 
we set nearest neighbors number $k$ as 10, both removal ratio $P_{rm}$ and clustering fraction $P_{in}$ as $10\%$, opponent penalty coefficient $\delta$ as $1.1$. The experiment details of supervised finetuning are listed in the Appendix~\ref{app:app_exp_setup}.

\subsection{Overall Peformance Comparison}
\label{sec:bench_result}
The average performance and standard deviation from three independent runs are presented in Tab.~\ref{tab:benchmarks}. Under a low sampling ratio of 0.5\% for CIFAR10, our method is slightly outperformed by ActiveFT. This can be attributed to the more stable model training aided by core point selection at extremely low budgets, whereas boundary samples tend to introduce greater instability and perturbation. However, as the data volume increases, the importance of constructing boundaries becomes more evident. In this regard, our approach, BiLAF, demonstrates a significant advantage across most scenarios, substantially outperforming competing models. Notably, on CIFAR100, our model consistently shows an improvement of approximately 3\% over the currently best-performing model, ActiveFT. Similarly, on ImageNet, BiLAF achieves a stable enhancement of about 1\%. These marked improvements highlight the efficacy of our BiLAF method.

\subsection{Qualitative Visualization}
\label{sec:analysis}
\begin{figure}[t!]
    \centering
    \includegraphics[width=0.49\linewidth]{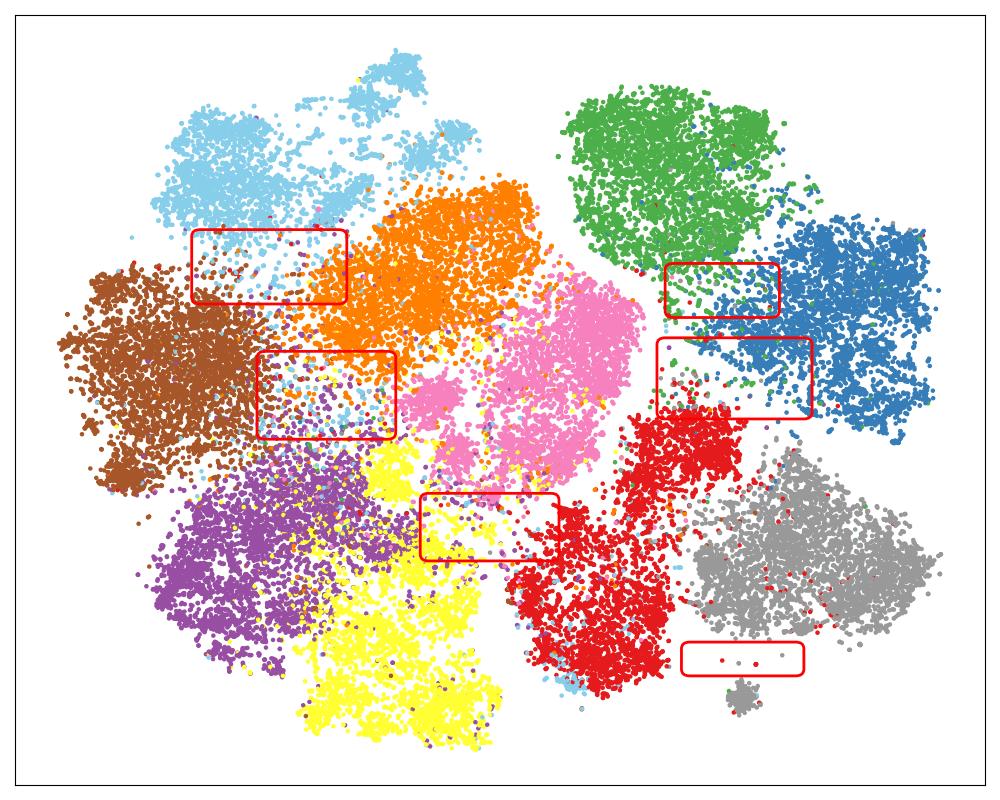}
    \includegraphics[width=0.49\linewidth]{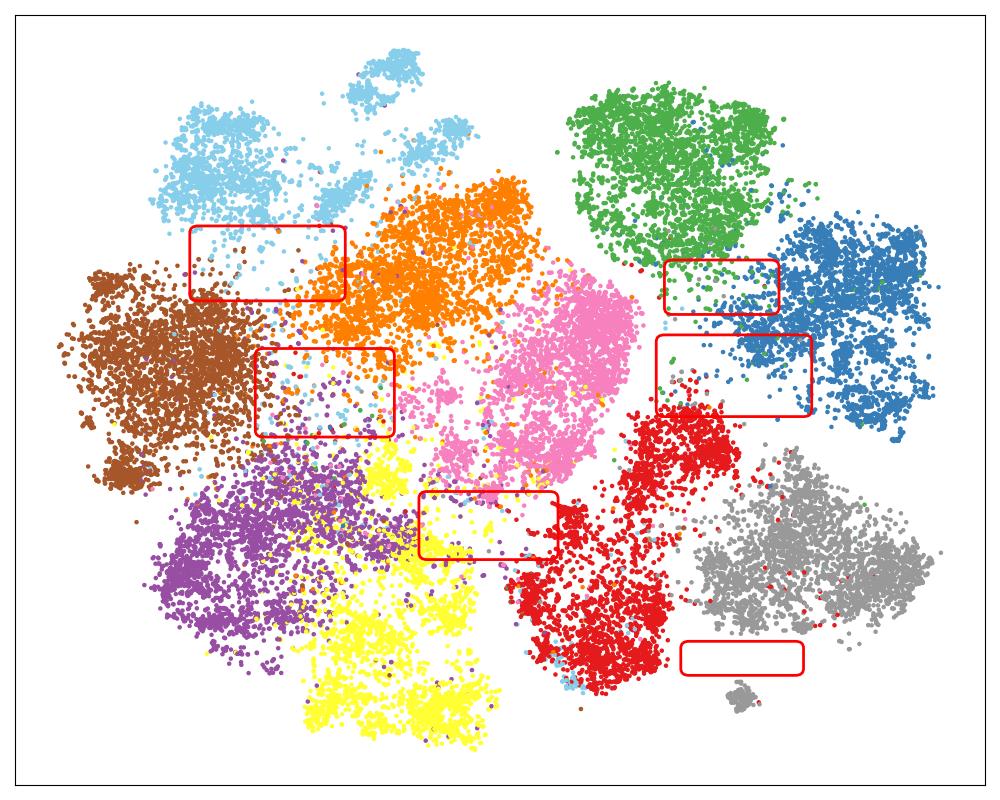}
    \caption{\textbf{Denoising  Visualization using tSNE Embeddings on CIFAR10} The left figure illustrates the initial samples, while the right figure depicts the scenario after denoising 30\% of the samples. The red bounding boxes highlight several conspicuous changes.}
    \label{fig:denoise}
\end{figure}

\begin{figure}[tb!]
    \centering
    \includegraphics[width=0.9\linewidth]{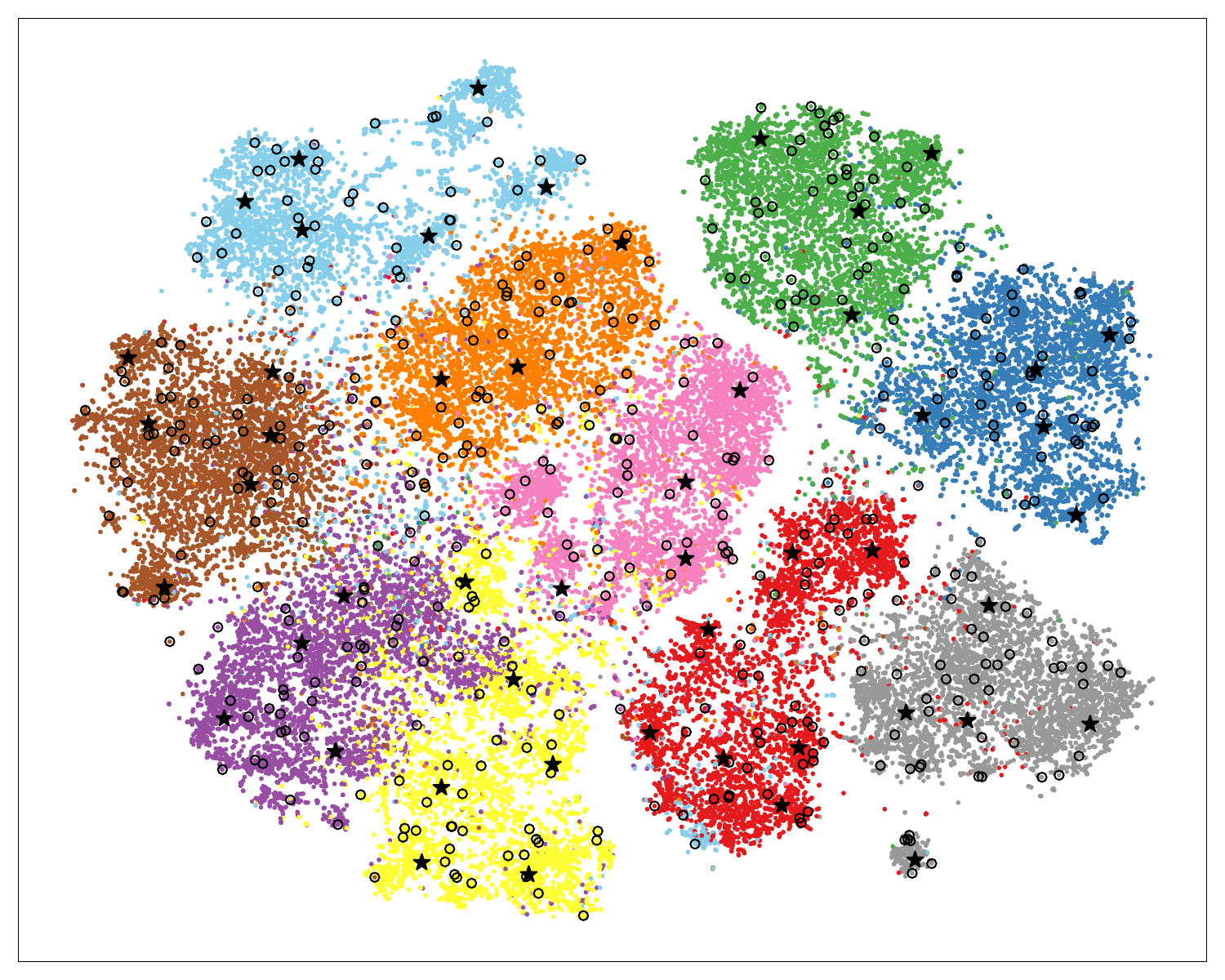}
    \caption{\textbf{tSNE Embeddings on CIFAR10 with 1\% annotation budget of our BiLAF method.} Pentagrams represent the selected core samples, while circles denote the chosen boundary samples.}
    \label{fig:samples_our}
\end{figure}

\paragraph{Denoising Visualization.} Fig.~\ref{fig:denoise} showcases the effects of the denoising process. For clarity, we choose the 30\% denoising intensity. It is evident that several isolated scatter points have been successfully eliminated. More detailed process can be found in the Appendix~\ref{app:app_visual}.

\paragraph{Selected Samples Visualization.} Fig.~\ref{fig:samples_our}
illustrates the sample selection process of our method. Firstly, we identify the central samples represented by pentagrams, and then expand to the boundary points denoted by circles from each center. Our method focuses on the boundaries between two categories rather than purely fitting the entire distribution. This approach allows for the selection of more meaningful points. 
For a more detailed comparison with other methods, please refer to the Appendix~\ref{app:app_visual}.

\subsection{Ablation Study}
\label{sec:ablation}

\begin{table*}[tb!]
    \centering
    \caption{\textbf{Ablation on CIFAR100.} In denoising process,``DG", ``DB" and ``IDC" indicate the basic distance-guide method,  density-based method and iterative density-based clustering method. In selection criterion,``BD", ``BS" indicate the basic distance metric and the boundary score metric. 
    In selection process,``OS", ``ISR", ``OP" indicate selecting the top samples in one shot, iterative selection and removal and whether to use opponent penalty. BiLAF represents the complete implementation and we explore the influence of three designs separately.}
    \vspace{-5pt}
    \resizebox{0.9\textwidth}{!}{
    \begin{tabular}{c|ccc|cc|ccc||cccc}
      \toprule[1.0pt]
       \multirow{2}{*}{\textbf{ID}} &
       \multicolumn{3}{c|}{\textbf{Denoising Process}} &
       \multicolumn{2}{c|}{\textbf{Selection Criterion}} &
       \multicolumn{3}{c||}{\textbf{Selection Process}} & \multicolumn{4}{c}{\textbf{Annotation budget}}   \\ 
        & DG & DB & IDC & \quad BD &  BS & OS &\  ISR & OP & $1\%$ & $2\%$ & $5\%$ & $10\%$   \\ 
       \midrule[1pt]
        BiLAF & - & - & \checkmark & - & \checkmark & - & \checkmark & \checkmark & \textbf{31.82} & \textbf{43.48} & \textbf{62.75} & \textbf{73.67} \\
        \midrule[0.8pt]
        1 & - & \checkmark & - & - & \checkmark & - & \checkmark & \checkmark & 31.10 \minus{0.72} & 41.06 \minus{2.42} & 61.60 \minus{1.15} & 73.39 \minus{0.28} \\
        2 & \checkmark & - & - & - & \checkmark & - & \checkmark & \checkmark & 29.27 \minus{2.55} & 36.42 \minus{7.06} & 61.16 \minus{1.59} & 72.87 \minus{0.80} \\
        3 & - & - & - & - & \checkmark & - & \checkmark & \checkmark & 28.88 \minus{2.94} & 36.78 \minus{6.70} & 60.83 \minus{1.92} & 72.76 \minus{0.91} \\
        \midrule[0.6pt]
        4 & - & - & \checkmark & \checkmark & - & - & \checkmark & \checkmark & 27.94 \minus{3.88} & 33.76 \minus{9.72} & 54.03 \minus{8.72} & 71.17 \minus{2.50} \\
        \midrule[0.6pt]
        5 & - & - & \checkmark & - & \checkmark & - & \checkmark & - & \textbf{32.13} \plus{0.31} & 43.09 \minus{0.39} & 62.07 \minus{0.68} & 72.92 \minus{0.75} \\
        6 & - & - & \checkmark & - & \checkmark & \checkmark & - & - & 30.52 \minus{1.30} & 42.13 \minus{1.35} & 60.24 \minus{2.51} & 69.66 \minus{4.01} \\
        
      \bottomrule[1.0pt]
    \end{tabular}
    }
    \label{tab:metric_influ}
\end{table*}

\paragraph{Effectiveness of Designs.}
Tab.~\ref{tab:metric_influ} shows the contributions of all the proposed components described in Sec.~\ref{sec:boundart_selction} to the model performance. Our framework is principally structured around three components: Denoising Process, Selection Criterion, and Selection Process. IDs 1 to 3 assess the impact of the Denoising Process design, comparing the use of iterative density-based clustering  with other denoising strategies detailed in Appendix~\ref{app:denoise} , or even the absence of denoising. We observe that performance degradation is substantial with a smaller budget, but narrows as the budget increases. This suggests that when fewer data are selected, the detrimental impact of noise samples is pronounced; however, as the data volume expands, the deficiencies are compensated, highlighting the effectiveness of our denoising approach. ID 4 illustrates the impact of our Selection Criterion design. Without employing the Boundary Score metric, models struggle to accurately identify potential boundary samples and are adversely affected by irrelevant marginal samples, significantly impairing performance. IDs 5 and 6 demonstrate the effects of our Selection Process design. The opponent penalty is a valuable addition, yet its removal does not critically affect performance, 
and may even result in improvements at 1\% data volume. 
This could be due to the blurred boundaries between certain categories that need greater focus. However, iterative selection and removal prove crucial, especially with increasing data volume, as they considerably enhance the diversity of boundary samples selected. As an one-go approach, the increased data volume leads to an accumulation of redundant samples, causing wastage and performance degradation.

\paragraph{Core Samples Selection Number.}
Without the ground truth labels, it is not always possible to ensure that each selected pseudo-class center represents a distinct category in the core sample selection stage of our BiLAF method. 
Tab.~\ref{tab:Core_number} illustrates the effect of choosing different numbers of core samples on the accuracy under various annotation budget within the CIFAR100 dataset. We discovered that performance is notably suboptimal when a smaller number of centers are selected; for instance, choosing 125 centers for 100 categories yielded poor results. This primarily stems from the inability of a limited number of centers to encompass all existing categories, leading to significant challenges in subsequent boundary point selection. However, once a sufficient number of centers to cover all categories is reached, performance tends to stabilize. Performance gains can be achieved by optimizing the ratio of center samples to boundary samples. For example, the best performance at 5\% data volume was observed with 375 centers, while at 10\% data volume, 500 centers proved to be optimal. This indicates that there is still room to enhance our method. However, for consistency, we chose to use the same number of centers  for all experiments.

\begin{table}[tb!]
    \centering
    \caption{\textbf{Ablation for Core Samples Numbers on CIFAR100.} 
    }
    \vspace{-5pt}
    \resizebox{0.9\linewidth}{!}{%
    \begin{tabular}{c|cccc}
        \toprule
            \multirow{2}{*}{\textbf{Budget}}  & \multicolumn{4}{c}{\textbf{Core Ratio / Core Number}}    \\ 

            & 0.25\% / 125 & 0.50\% / 250 & 0.75\% / 375 & 1.00\% / 500 \\

            \midrule
            $1\%$ & 21.51  & \textbf{31.82} & 28.37 &  27.24 \\
            $2\%$ & 36.64  & \textbf{43.48} & 42.68 &  42.18 \\
            $5\%$ & 59.20  & 62.75 & \textbf{63.32} &  62.46\\
            $10\%$ & 71.86  & 73.67 & 73.58 & \textbf{74.32} \\
            \bottomrule
    \end{tabular}%
    } 
    \label{tab:Core_number}
\end{table}

\begin{table}[tb!]
    \centering
    \caption{\textbf{Ablation for Core Selection Method on CIFAR100.} 
    }
    \vspace{-5pt}
    \resizebox{0.8\linewidth}{!}{%
    \begin{tabular}{c|cccc}
        \toprule
            \multirow{2}{*}{\textbf{Budget}}  & \multicolumn{4}{c}{\textbf{Core Selection Method}}    \\ 
            & Random & FDS & K-Means & ActiveFT \\
            \midrule
            $1\%$ &  25.58 & 20.69 & 28.80 & \textbf{31.82} \\
            $2\%$ & 36.22  & 33.22 & 41.17 &  \textbf{43.48}\\
            $5\%$ & 60.68  & 60.13 & 62.39 &  \textbf{62.75}\\
            $10\%$ & 72.84  & 71.05 & \textbf{74.27} & 73.67 \\
            \bottomrule
    \end{tabular}%
    } 
    \label{tab:Core_method}
    \vspace{-5pt}
\end{table}

\paragraph{Core Samples Selection Method.} 
In core samples selection, we defaulted to ActiveFT as our method of choice. However, there are numerous existing methods, such as Random, FDS, and K-Means. 
Tab.~\ref{tab:Core_method} presents the model performance based on boundary selection using different pseudo-class centers.
We observe that the accuracy of the BiLAF framework is positively correlated with the quality of the model used for selecting pseudo-class centers. This is because the mis-selection of centers can introduce significant bias, greatly affecting subsequent sample selection. Among these, ActiveFT performs the best, while the traditional K-Means method also demonstrates strong performance, confirming the universality of our framework when based on well-defined centers.

\begin{figure}[t!]
    \centering
    \includegraphics[width=0.49\linewidth]{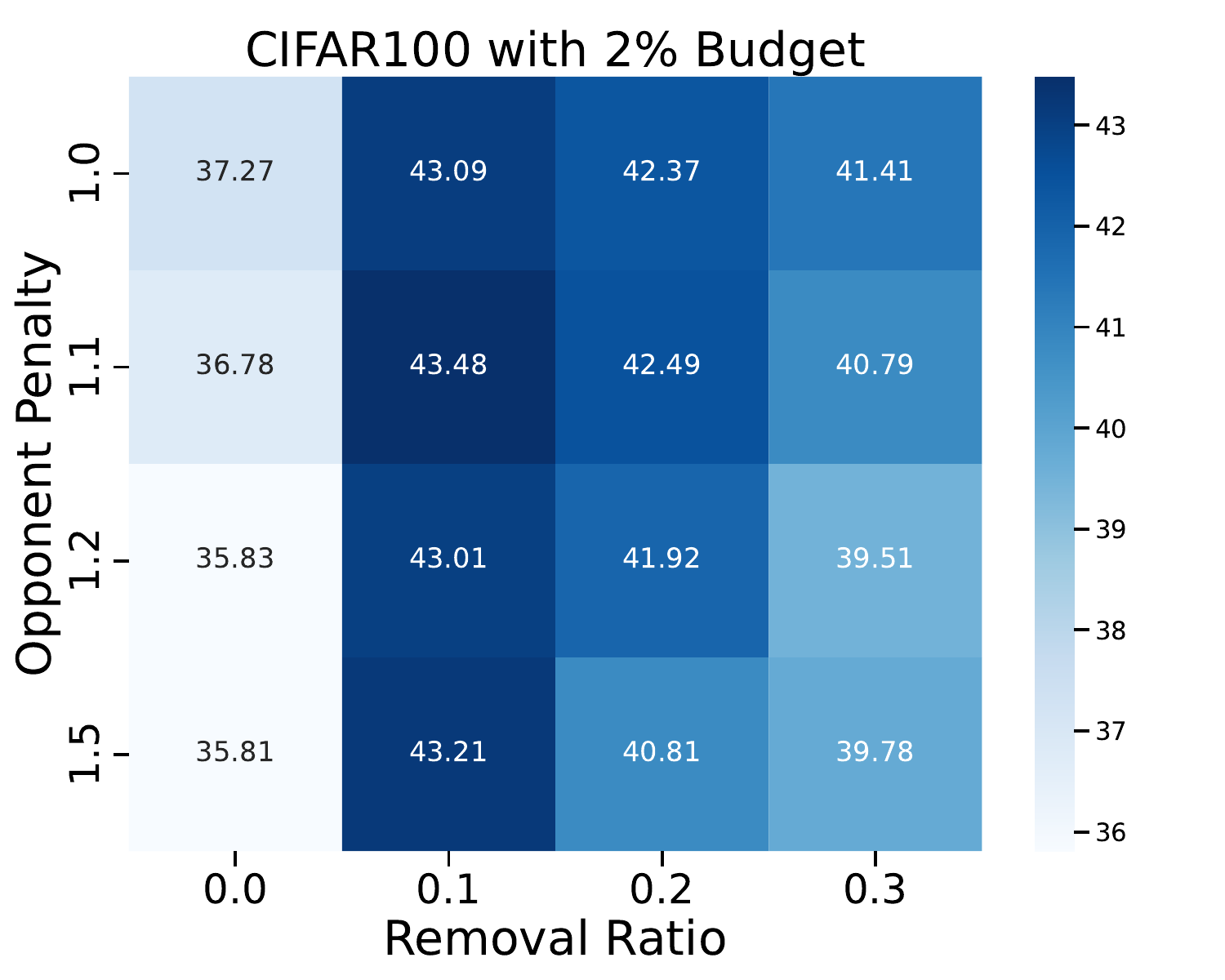}
    \includegraphics[width=0.49\linewidth]{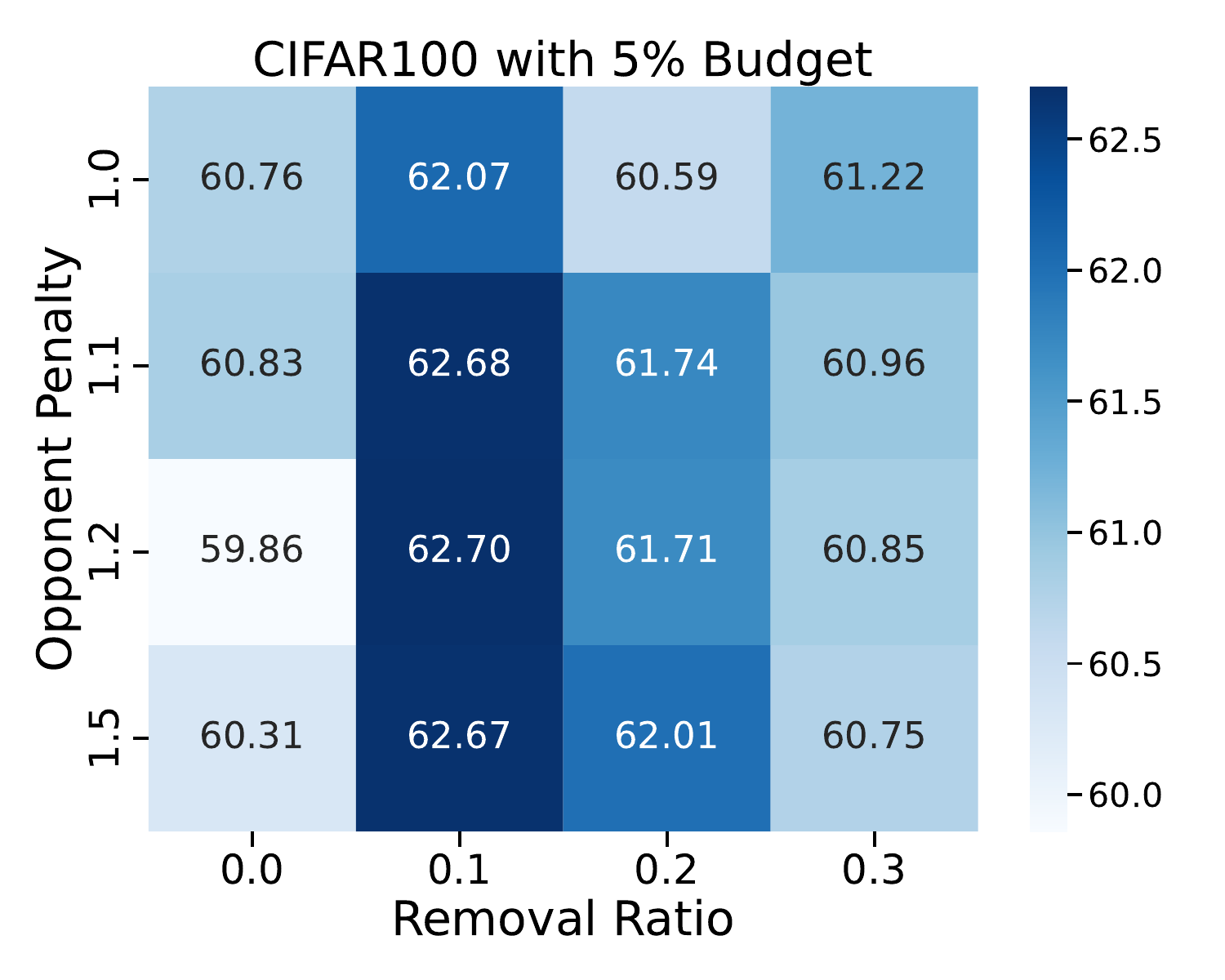}
    \vspace{-5pt}
    \caption{\textbf{The Hyperparameter Influence on CIFAR100.} }
    \label{fig:hyper_heatmap}
    \vspace{-5pt}
\end{figure}


\paragraph{Hyperparameter Influence.} Our primary handcrafted parameters include the removal rate of noise points during the denoising process, and the opponent penalty applied during the selection process for penalizing the same class boundaries. Fig.~\ref{fig:hyper_heatmap} explores the impact of varying removal rates and opponent penalties on the performance of our model on CIFAR100, with annotation budgets of 2\% and 5\%. We observe that the removal rate has a significant impact. If set too low, it leads to an excess of noise; if set too high, it results in a lack of boundary sample points when the budget is increased, thereby affecting performance. In contrast, the impact of opponent penalty is relatively minor, though it can enhance model performance to a certain extent.

\section{Conclusion}

In this paper, we highlight the significance of active finetuning tasks and critically examine existing methods, noting their limited consideration of uncertainty, which is especially pronounced when the pretraining-finetune paradigm is adopted. We introduce an innovative solution: the Bi-Level Active Finetuning Framework (BiLAF). This framework not only ensures diversity in the selection of central points but also emphasizes boundary samples with higher uncertainty. BiLAF effectively integrates existing core sample selection models and introduces a novel strategy for boundary sample selection in unsupervised scenarios. Our extensive empirical studies validate the effectiveness of BiLAF, showcasing its ability to enhance predictive performance.  Through comparative experiments, we also explore new avenues, such as finding the optimal balance between central and boundary points. We are confident that our work offers valuable insights into Active Finetuning and will serve as a catalyst for further research in this evolving field.

\clearpage

\bibliographystyle{named}
\bibliography{ijcai24}

\clearpage

\appendix

\section{Details of ActiveFT Method}
\label{app:app_active}
ActiveFT~\cite{xie2023active} selects the most useful data samples in the feature space of the pretrained model under the guidance of two basic intuitions: \textit{1) bringing close the distributions between the selected subset $\mathcal{P}^u_{\mathcal{S}}$ and the original pool $\mathcal{P}^u$. 2) maintaining the diversity of $\mathcal{P}^u_{\mathcal{S}}$.} Formally, the goal is to find the optimal selection strategy $\mathcal{S}$ as follows.
\begin{equation}
\mathcal{S}_{opt}=\arg\min_{\mathcal{S}}D(p_{f_u},p_{f_{\mathcal{S}}})-\lambda R(\mathcal{F}_{\mathcal{S}}^u)
\end{equation}
where $D(\cdot,\cdot)$ is some distance metric between distributions $p_{f_u}$ of $\mathcal{P}^u$ and  $p_{f_{\mathcal{S}}}$ of $\mathcal{P}^u_{\mathcal{S}}$, $R(\cdot)$ is to measure the diversity of a set, and $\lambda$ is a scale to balance these two terms.

Due to the difficulty in directly optimizing the discrete selection strategy $\mathcal{S}$, $p_{f_{\mathcal{S}}}$ is alternatively modeled with $p_{\theta_{\mathcal{S}}}$, where $\theta_{\mathcal{S}}=\{\theta_{\mathcal{S}}^j\}_{j\in[K]}$ are the continuous parameters and $K$ is the budget size of core samples. 
Each $\theta_{\mathcal{S}}^j$ after optimization corresponds to the feature of a selected sample $\mathbf{f}_{s_j}$. 
We would find $\mathbf{f}_{s_j}$ closest to each $\theta_{\mathcal{S}}^j$ after optimization to determine the selection strategy $\mathcal{S}$. The goal of this continuous optimization is written in Eq.~\ref{eq:goal}.

\begin{equation}
    \theta_{\mathcal{S},opt}=\arg\min_{\theta_{\mathcal{S}}}D(p_{f_u},p_{\theta_{\mathcal{S}}})-\lambda R(\theta_{\mathcal{S}})\ \ \ s.t. ||\theta_{\mathcal{S}}^j||_2=1
    \label{eq:goal}
\end{equation}

Following the mathematical derivation in \cite{xie2023active}, Eq.~\ref{eq:goal} can be solved by optimizing the following loss function through gradient descent.
\begin{equation}
\label{eq:loss}
        \resizebox{\linewidth}{!}{
        \begin{math}
    \begin{aligned} 
        L&=D(p_{f_u},p_{\theta_{\mathcal{S}}})-\lambda\cdot  R(\theta_{\mathcal{S}})\\
        &=-\underset{\mathbf{f}_i\in \mathcal{F}^u}{E}\left[sim(\mathbf{f}_i,\theta_{\mathcal{S}}^{c_i})/\tau\right]+\underset{j\in[B]}{E}\left[\log\sum_{k\neq j,k\in[B]}\exp\left(sim(\theta_{\mathcal{S}}^j,\theta_{\mathcal{S}}^k)/\tau\right)\right]
    \end{aligned}
    \end{math}
    }
\end{equation}
where $sim(\cdot,\cdot)$ is a cosine similarity between normalized features, the temperature $\tau=0.07$ and the balance weight $\lambda$ is empirically set as $1$.

After the optimization process, it finds features $\{\mathbf{f}_{s_j}\}_{j\in[K]}$ with the highest similarity to $\theta_{\mathcal{S}}^j$.
\begin{equation}
    \mathbf{f}_{s_j}=\arg\max_{\mathbf{f}_k\in\mathcal{F}^u}sim(\mathbf{f}_k,\theta_{\mathcal{S}}^j)
    \label{eq:sim}
\end{equation}
The corresponding data samples $\{\mathbf{x}_{s_j}\}_{j\in[K]}$ are selected as the subset $\mathcal{P}_{\mathcal{S}}^u$ with selection strategy $\mathcal{S}=\{s_j\}_{j\in[K]}$. 

\section{Denoising Algorithm}
\label{app:denoise}
We propose three progressive methods as a starting point to address this challenge of denoising, including the distance-guide method,  density-based method, iterative density-based clustering (IDC). The IDC method is elaborated in detail within the main body of the paper. In this section of the appendix, we introduce two additional methods.

\textbf{Distance-guide method} removes the top 
$N_{i,rm}$
samples furthest from the current class center based on their distance. This approach is straightforward and simple; however, it has limitations, particularly in cases where the class distribution varies significantly in different directions, resembling an elliptical distribution, for instance. In such scenarios, this rudimentary method may inadvertently remove the farthest cluster of samples rather than the actual noise samples.

\textbf{Density-based method} operates by removing the top $N_{i,rm}$ points with the largest density distance from a class. The underlying logic is that noise points are generally farther away from their neighboring points. Thus, using the distance to nearby points as an auxiliary measure to identify noise is reasonable. The density distance is defined in Eq.~\ref{equ:density} in the main body.

\begin{figure*}[tb!]
    \centering
    \begin{subfigure}{0.4\linewidth}
        \centering
    \includegraphics[width=\textwidth]{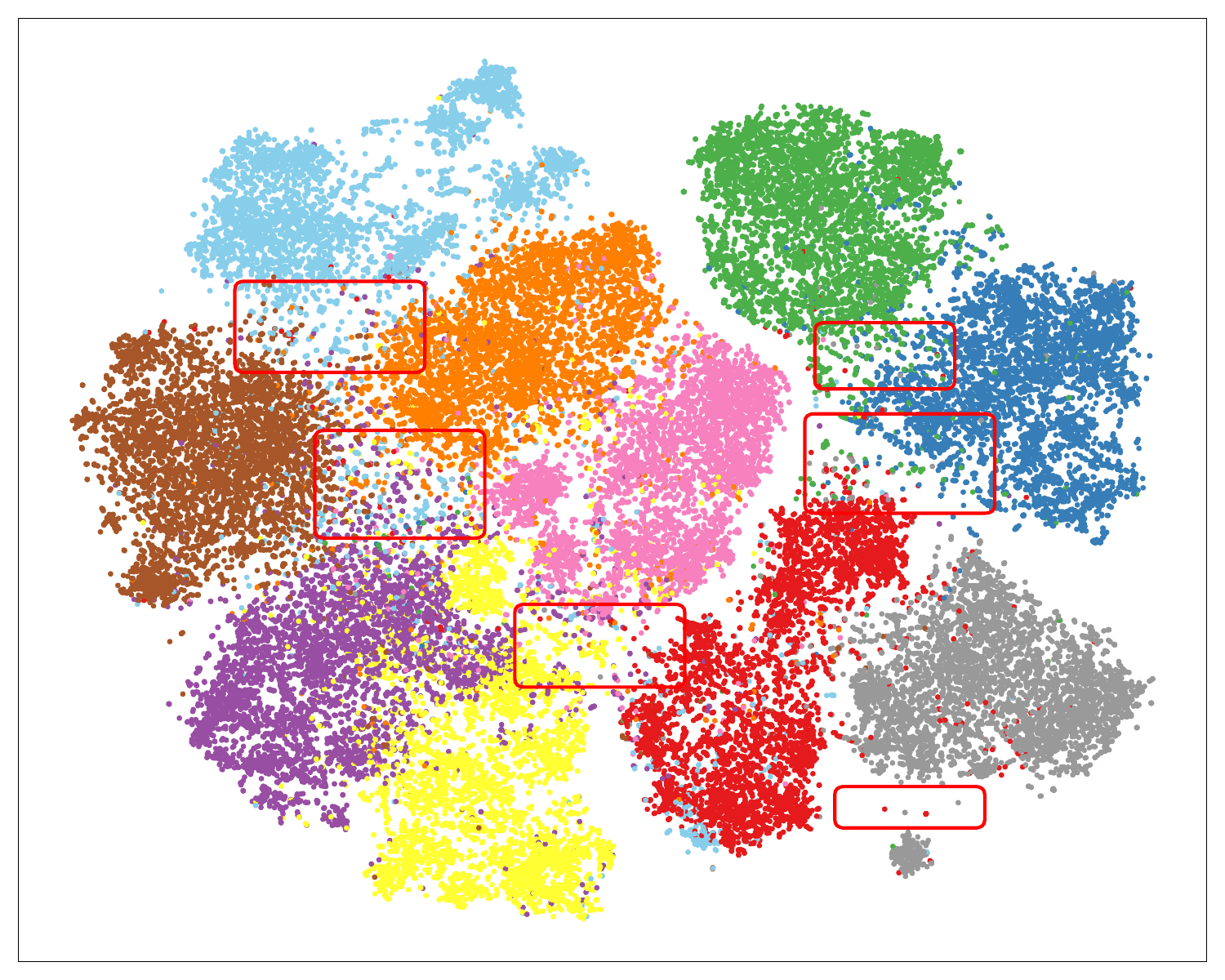}
        \caption{Denoising Ratio $P_{rm} = 0\%$}
    \end{subfigure}
    \begin{subfigure}{0.4\linewidth}
        \centering
    \includegraphics[width=\textwidth]{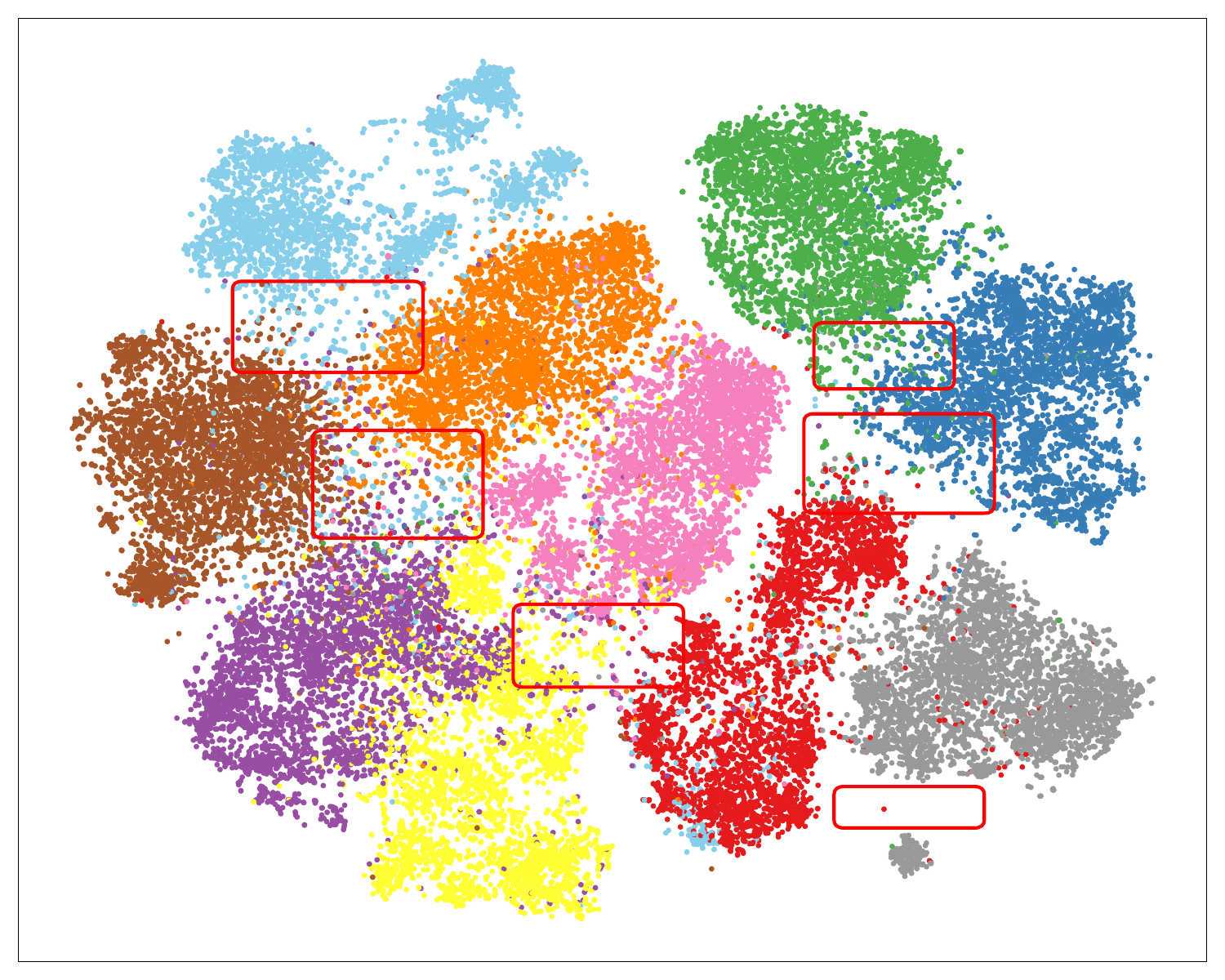}
        \caption{Denoising Ratio $P_{rm} = 10\%$}
    \end{subfigure}
    \begin{subfigure}{0.4\linewidth}
        \centering
    \includegraphics[width=\textwidth]{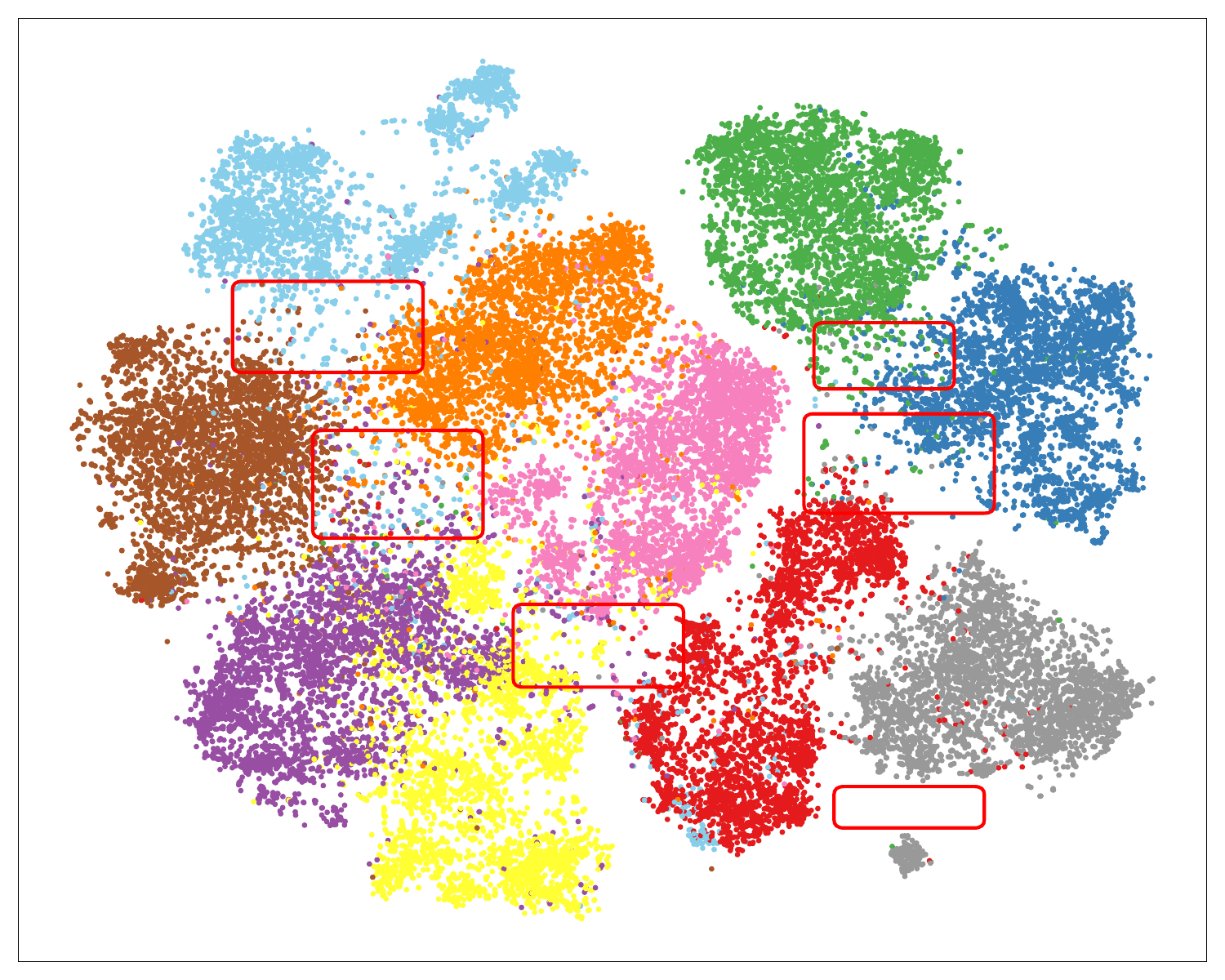}
        \caption{Denoising Ratio $P_{rm} = 20\%$}
    \end{subfigure}
    \begin{subfigure}{0.4\linewidth}
        \centering
    \includegraphics[width=\textwidth]{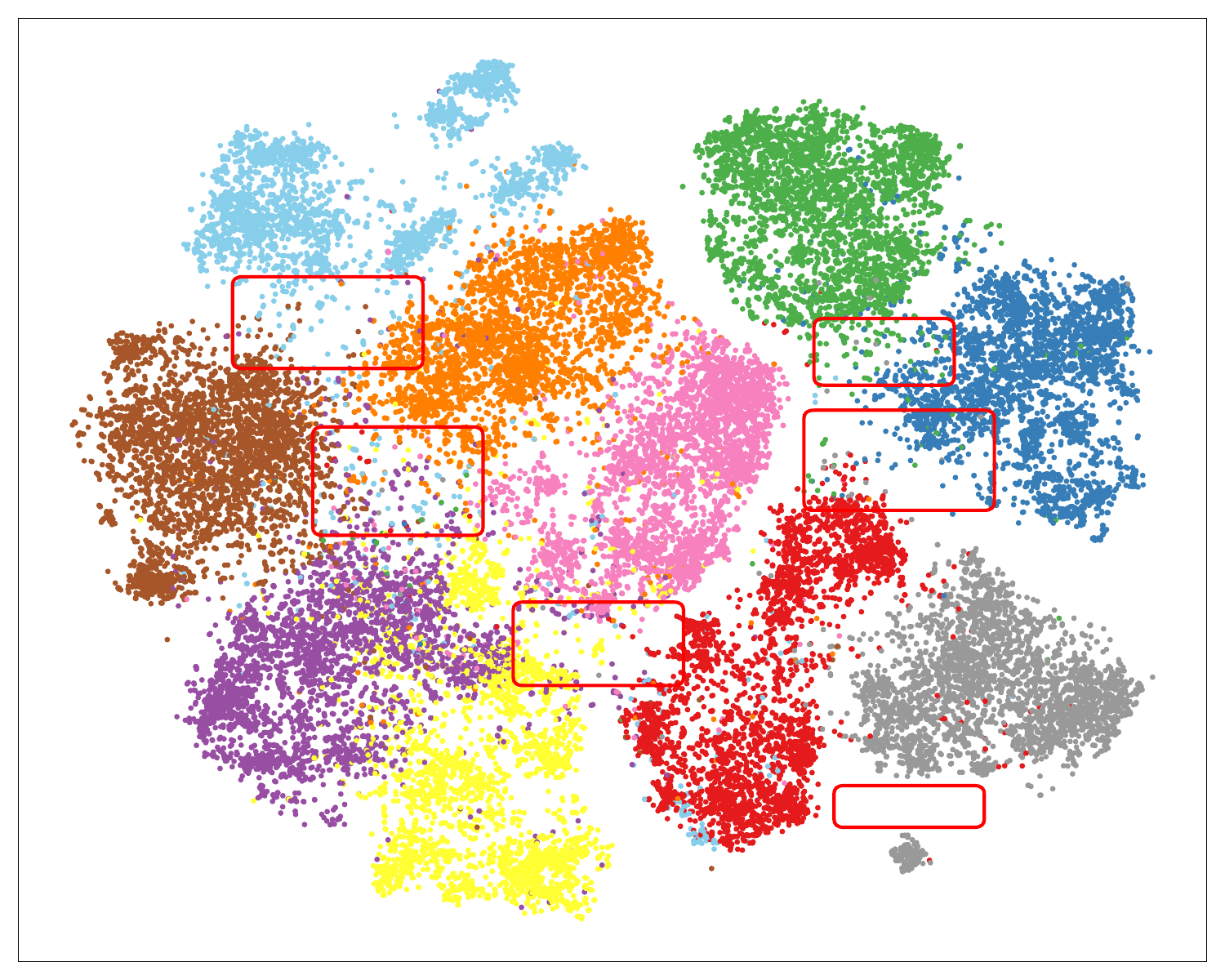}
        \caption{Denoising Ratio $P_{rm} = 30\%$}
    \end{subfigure}
    \caption{\textbf{Denoising Strength Visualization using tSNE Embeddings of CIFAR10}.}
    \label{fig:app_denoise}
\end{figure*}

\section{Experiment Setup}
\label{app:app_exp_setup}
Due to space constraints in the main part, we provide a comprehensive overview of the experiment setup in this section.

\textbf{Datasets and Evaluation Metrics.} Our approach is evaluated using three widely recognized datasets: CIFAR10, CIFAR100 \cite{krizhevsky2009learning}, and ImageNet-1k \cite{russakovsky2015imagenet}. Both CIFAR10 and CIFAR100 contain 60,000 images with resolutions of 32x32, but they differ in classification complexity, offering 10 and 100 categories, respectively. Each comprises 50,000 images for training and 10,000 for testing. The large-scale dataset ImageNet-1k includes 1,000 categories and a total of 1,281,167 training images along with 50,000 validation images. The raw training data serve as the unlabeled pool $\mathcal{P}^u$ from which selections are made. We employ the \textit{Top-1 Accuracy} metric for performance evaluation.

\textbf{Baselines.} We compare our approach with three heuristic baselines, four active learning methods and the well-designed active finetuning method ActiveFT~\cite{xie2023active}. 

Three heuristic baselines are listed as follows: 1. \textit{Random}: The selection of samples for annotation is entirely stochastic. 2. \textit{FDS}: The method is also known as the K-Center-Greedy algorithm, which selects the next sample feature that is farthest from the current selections. As proven in~\cite{sener2017active}, it minimizes the expected loss over the entire pool and the selected subset. 
3. \textit{K-Means}: We implement the K-Means method on the feature pool $\mathcal{F}^u$ and select samples nearest to the centroids, where the number $K$ equals to the budget size $B$.

Four active learning methods contains CoreSet~\cite{sener2017active}, VAAL~\cite{sinha2019variational}, LearnLoss~\cite{yoo2019learning}, TA-VAAL~\cite{kim2021task}, and ALFA-Mix~\cite{parvaneh2022active}. These methods has been modified and applied specifically for active finetuning tasks following the instructions in ~\cite{xie2023active}.  These approaches utilize a batch-selection strategy for data sampling. Initially, the model is trained on a randomly chosen initial dataset. Subsequently, it employs this trained model to pick a batch of images from the training set. The model is then retrained using the cumulatively selected samples. The chosen active learning techniques encompass strategies based on both diversity and uncertainty within the active learning domain, serving as the baselines in the active finetuning field.

In this scenario, ActiveFT~\cite{xie2023active} emerges as the strongest baseline due to its tailored design for the task. This approach generates a representative subset from the unlabeled pool by aligning its distribution with the entire dataset and enhances diversity through the optimization of parametric models within a continuous space.

\textbf{Implementation Details.} In the unsupervised pretraining phase, we use DeiT-Small~ \cite{touvron2021training} model, pretrained using the DINO~\cite{caron2021emerging} framework on ImageNet-1k, due to its recognized efficiency and popularity.
For all three datasets, we resize images to $224\times224$ consistent with the pretraining for both data selection and supervised finetuning. 
In core samples selection stage, we utilize ActiveFT and optimize the parameters $\theta_{\mathcal{S}}$ using the Adam \cite{kingma2014adam} optimizer (learning rate 1e-3) until convergence. We set the core number $K$ as $50(0.1\%), 250(0.5\%), 6405(0.5\%)$  for CIFAR10, CIFAR100 and ImageNet separately. In the boundary samples selection stage, 
we set nearest neighbors number $k$ as 10, both removal ratio $P_{rm}$ and clustering fraction $P_{in}$ as $10\%$, opponent penalty coefficient $\delta$ as $1.1$. In the supervised finetuning phase, we finetune DeiT-Small model following the setting in~\cite{xie2023active}.  We finetune the models using the SGD optimizer with learning rate as $3e$-$3$, weight decay as $1e$-$4$ and momentum as $0.9$. We employ cosine learning rate decay with a batch size of 256 distributed across two GPUs. The models are finetuned for 1000 epochs on all datasets with different sampling ratios, except for ImageNet with sampling ratio 5\%, where we finetune for 300 epochs.

\section{Qualitative Visualization}
\label{app:app_visual}
\paragraph{Denoising Visualization.} Fig.~\ref{fig:app_denoise}  illustrates the impact of variations in the removal rate $P_{rm}$ during the denoising process on the retained samples. The red bounding boxes highlight regions with significant changes. These areas include obvious outliers and zones of confusion where multiple class intermingle. As the removal rate increases, the number of samples in these areas tends to decrease gradually, thereby reducing their influence on subsequent boundary point selection, while the relatively dense boundaries are often preserved. However, this represents a trade-off as some important samples might also be removed. Therefore, we have conducted a quantitative analysis in the main text to address this concern.

\begin{figure*}[tb!]
    \centering
    \begin{subfigure}{0.4\linewidth}
        \centering
    \includegraphics[width=\textwidth]{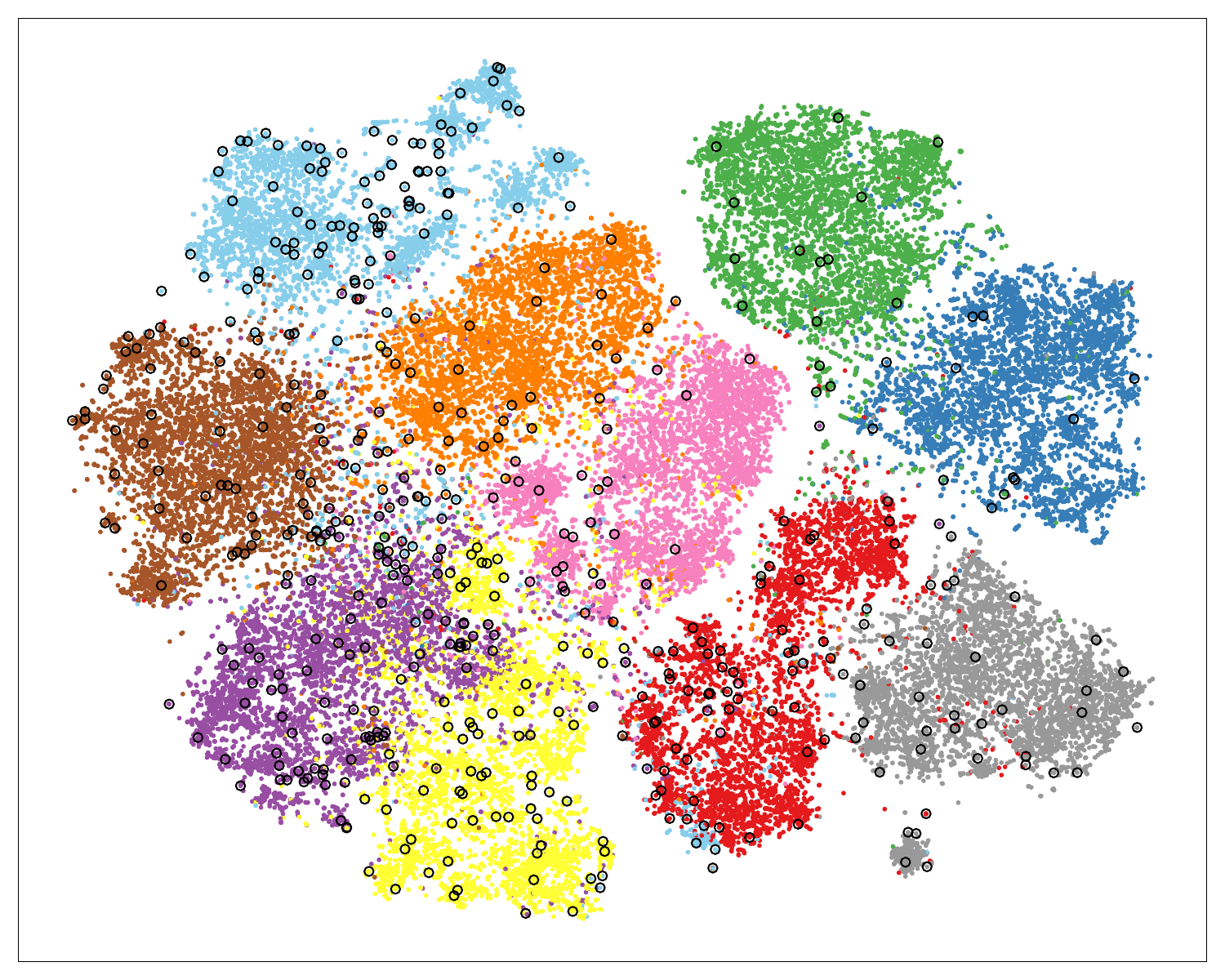}
        \caption{FDS}
    \end{subfigure}
    \begin{subfigure}{0.4\linewidth}
        \centering
    \includegraphics[width=\textwidth]{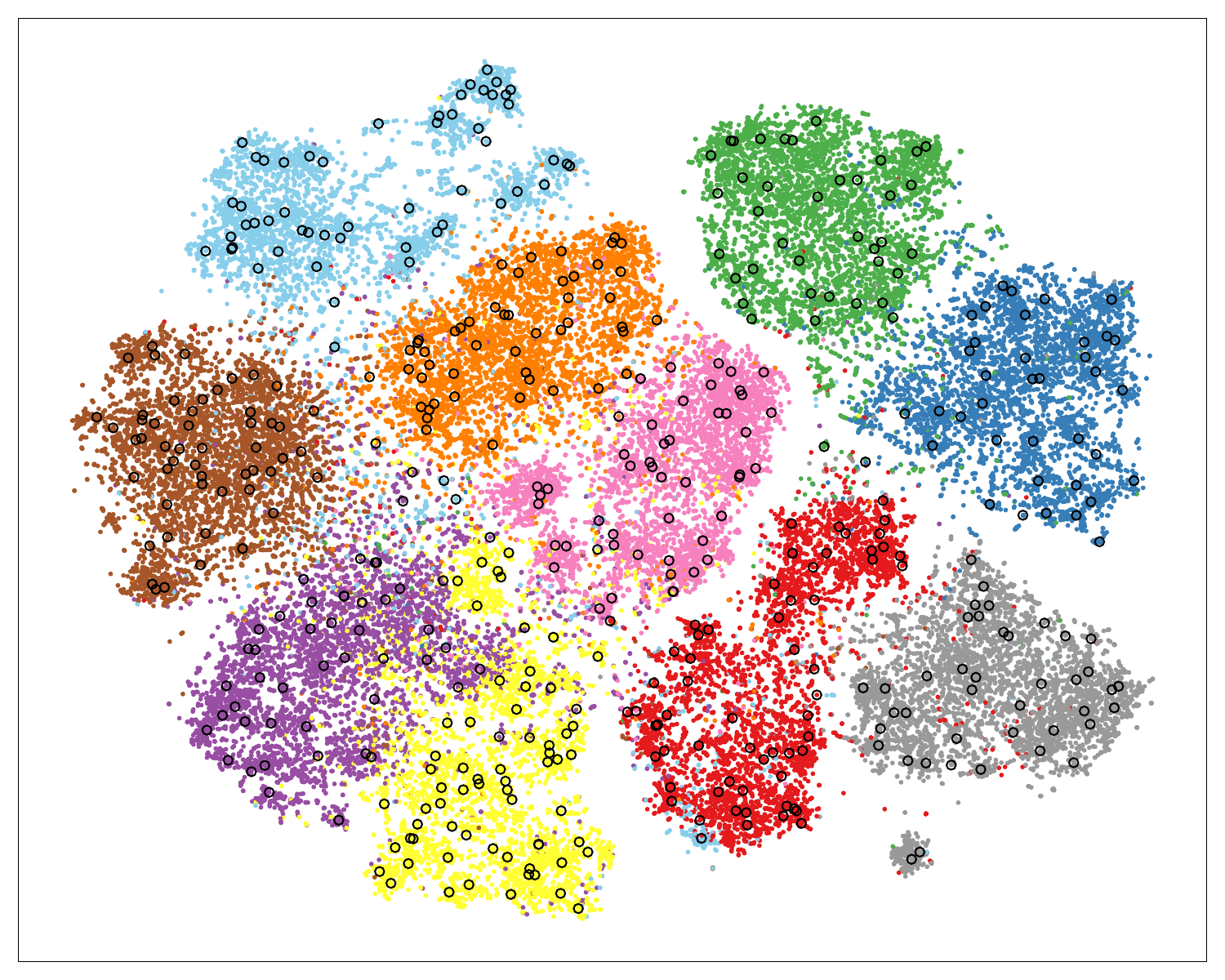}
        \caption{K-Means}
    \end{subfigure}
    \begin{subfigure}{0.4\linewidth}
        \centering
    \includegraphics[width=\textwidth]{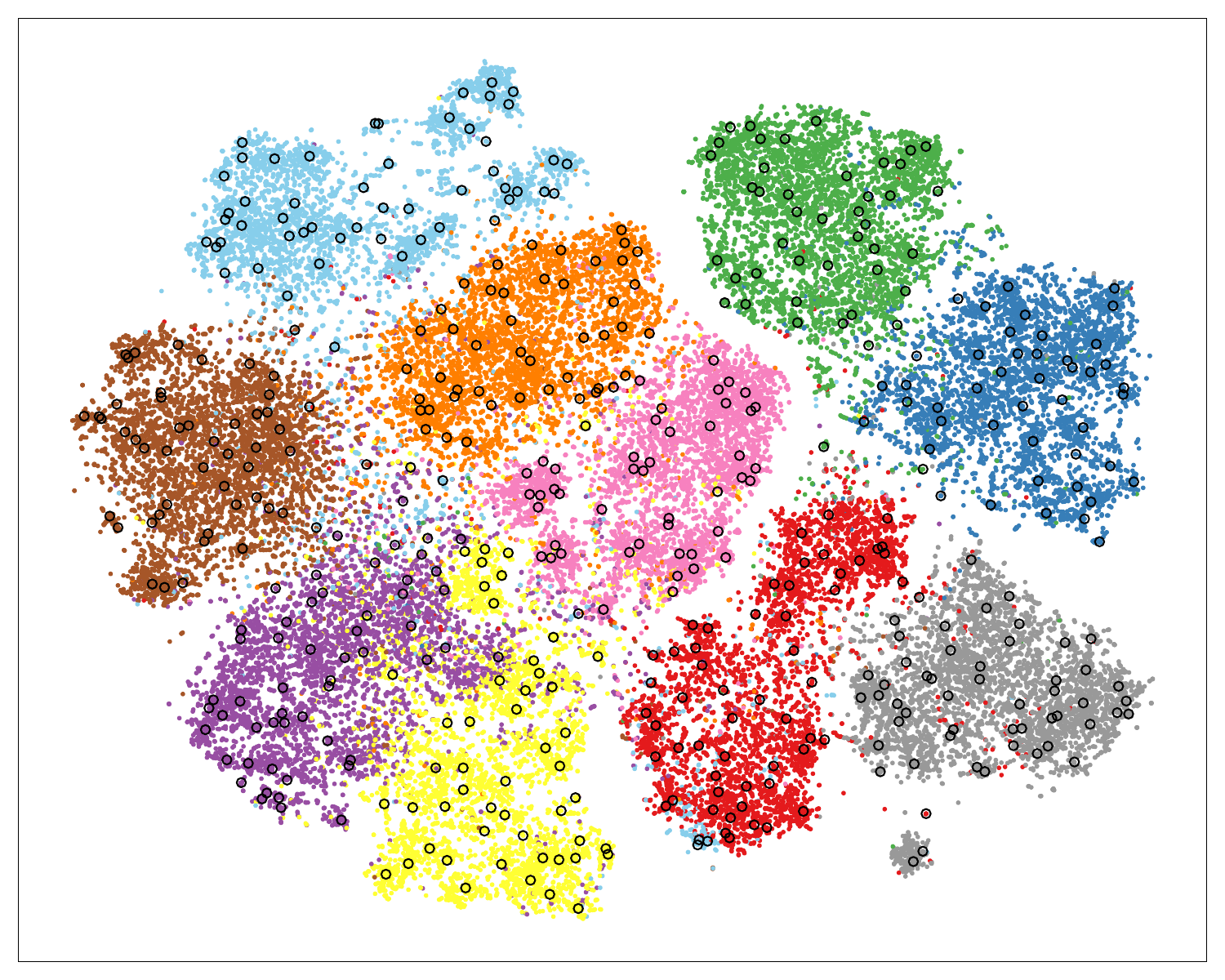}
        \caption{ActiveFT}
    \end{subfigure}
    \begin{subfigure}{0.4\linewidth}
        \centering
    \includegraphics[width=\textwidth]{figure/tsne_ours.pdf}
        \caption{BiLAF (ours)}
    \end{subfigure}
    \caption{\textbf{tSNE Embeddings on CIFAR10 with 1\% annotation budget of our BiLAF method and other methods.}}
    \label{fig:app_methods_vis}
\end{figure*}

\paragraph{Selected Samples Visualization.} Fig.~\ref{fig:app_methods_vis} displays the sample selection by different methods when annotating 1\% samples from the CIFAR100 dataset. It is evident that the FDS method has significant drawbacks, leading to an insufficient selection of sample samples from central classes in feature space, which greatly impedes the model's learning capability. Both the K-Means and ActiveFT methods focus on selecting central samples, differing in their optimization goals and processes. As seen in the Fig~\ref{fig:app_methods_vis}, both methods achieve their intended purpose, resulting in a relatively uniform selection of samples. Our method further refines this approach. Firstly, we identify the central samples, represented by pentagrams, and then expand to boundary samples, denoted by circles, from each center. Our strategy emphasizes the boundaries between categories, rather than conforming to the entire distribution. For example, considering the 'brown' sample class, our method selects fewer samples in its internal area, far from other categories, and focuses more on locating boundary samples. Extensive experimental results quantitatively demonstrate the substantial improvement our method brings to model performance, validating its effectiveness.

\end{document}